\documentclass[12pt]{article}
\usepackage{amsmath,amssymb}
\usepackage{times}
\usepackage{graphicx}
\usepackage{subcaption}
\usepackage{xcolor}   %
\usepackage{multirow}
\usepackage[style=apa, backend=biber]{biblatex}
\usepackage{rotating}
\usepackage{bbm}
\usepackage{latexsym}
\usepackage[title]{appendix}
\usepackage{hyperref}
\usepackage{array}

\usepackage{fullpage}

\newtheorem{Theorem}{Theorem}

\usepackage{authblk}

\title{Comparing Dynamical Models Through Diffeomorphic Vector Field Alignment}
\author[1]{Ruiqi Chen\thanks{Address correspondence to chen.ruiqi@wustl.edu}}
\author[2]{Giacomo Vedovati}
\author[3]{Todd Braver}
\author[2]{ShiNung Ching}
\affil[1]{Division of Biology and Biomedical Sciences, Washington University in St. Louis, \protect\\
St. Louis, MO, United States.}
\affil[2]{Department of Electrical and Systems Engineering, Washington University in St. Louis, \protect\\
St. Louis, MO, United States.}
\affil[3]{Department of Psychological and Brain Sciences, Washington University in St. Louis, \protect\\
St. Louis, MO, United States.}
\date{}

\begin{document}

\maketitle

\vspace{-2em}

\begin{abstract}

Dynamical systems models such as recurrent neural networks (RNNs) are increasingly popular in theoretical neuroscience as a vehicle for hypothesis-generation and data analysis.
Evaluating the dynamics in such models is key to understanding their learned generative mechanisms.
However, such evaluation is impeded by two major challenges:
First, comparison of learned dynamics across models is difficult because \textit{a priori} there is no enforced equivalence of their coordinate systems.
Second, identification of mechanistically important low-dimensional motifs (e.g., limit sets) is analytically intractable in high-dimensional nonlinear models such as RNNs.
Here, we propose a comprehensive framework to address these two issues, termed Diffeomorphic vector field alignment FOR learned Models (DFORM).
DFORM learns a nonlinear coordinate transformation between the state spaces of two dynamical systems, which aligns their trajectories in a maximally one-to-one manner.
In so doing, DFORM enables an assessment of whether a set of models exhibit topological equivalence, i.e., their dynamics are mechanistically similar despite differences in coordinate systems.
A byproduct of this method is a means to locate dynamical motifs on low-dimensional manifolds embedded within higher-dimensional systems.
We verified DFORM's ability to identify linear and nonlinear coordinate transformations using canonical topologically equivalent systems, RNNs, and systems related by nonlinear flows.
DFORM was also shown to provide a quantification of similarity between topologically distinct systems.
We then demonstrated that DFORM can locate important dynamical motifs including invariant manifolds and saddle limit sets within high-dimensional models.
Finally, using a set of RNN models trained on human functional magnetic resonance imaging (fMRI) recordings, we illustrated that DFORM can identify limit cycles from high-dimensional data-driven models, which agreed well with prior numerical analysis. 

\end{abstract}

\section{Introduction}

Recent development in theoretical neuroscience and scientific machine learning has leveraged the use of dynamical models as tools to characterize complex physical and computational processes \parencite{bruntonDiscoveringGoverningEquations2016, sussillo2014neural}. In neuroscience, dynamical models like recurrent neural networks (RNNs) are increasingly used as surrogates for brain networks to elucidate how the brain implements various functions \parencite{barak2017recurrent}. Dynamical models can be designed \parencite{wongRecurrentNetworkMechanism2006a}, trained to approximate experimental data \parencite{decoRestingStateFunctionalConnectivity2013, singhEstimationValidationIndividualized2020}, or most commonly, trained in a top-down fashion to perform cognitive tasks \parencite{yang2019task, sussillo2009generating}.
A key goal in the analysis of these models is to elucidate and compare the mechanisms -- the dynamics -- that they embed \parencite{maheswaranathan2019universality}. The analyses must be based on the topology and geometry of the vector field that defines the dynamics, sometimes referred to as the attractor landscape.  These landscapes, in turn, can embed low-dimensional manifolds along which neural activity evolves, a topic of considerable interest \parencite{langdonUnifyingPerspectiveNeural2023}.
However, there are at least two major challenges in such analyses.

First, vector fields may be deformed in rather arbitrary ways across models (moving equilibria around in state space, distorting the limit cycles, permuting the state variables, etc.), despite implementing qualitatively the same dynamics and functions. In particular, vector fields that emerge via stochastic optimization on tasks do not maintain fixed coordinate systems \textit{a priori}.
Coordinate systems can vary between sessions and subjects even in models fit to experimental data. For example, it is nearly impossible to empirically assign one-to-one correspondence between two sets of neurons recorded from two animals. Such heterogeneity in coordinate system makes it difficult to compare dynamics across models to reason about their shared and unique mechanisms.

Second, it is almost impossible to analytically locate the important features, such as limit sets, in high-dimensional nonlinear models (e.g., RNNs). Worse yet, while \emph{probable} limit sets (e.g., \emph{attractors}) can be found numerically through forward simulation of trajectories, \emph{improbable} limit sets (e.g., \emph{saddles}) cannot be found in this way, because the set of states that converge towards them has zero measure \parencite{abrahamDynamicsGeometryBehavior1992}. However, improbable limit sets hold a mechanistically important role in many computational models, such as mediating the transition between metastable brain states \parencite{robertsMetastableBrainWaves2019}, or implementing the winner-take-all mechanism in decision-making \parencite{wongRecurrentNetworkMechanism2006a}. A general way to identify such dynamical motifs will thus be crucial for reverse-engineering the mechanisms embedded within high-dimensional models.

In this paper, we developed an approach to address these two issues under a unifying framework termed Diffeomorphic vector field alignment FOR learned Models (DFORM). DFORM is a mathematical-computational technique designed to align the vector fields of two high-dimensional dynamical systems, thus enabling direct comparison of their geometry. DFORM can also align a high-dimensional system to a low-dimensional `template' to locate its important dynamical motifs.
To align two models with different coordinate systems, we appeal to the fundamental, dynamical systems theoretic notion of \emph{smooth orbital equivalence}. Two systems are smoothly orbitally equivalent if there exists a \emph{diffeomorphism} (a smooth coordinate transformation) between their respective phase spaces that matches the orbits (trajectories) of one system to those of the other in a one-to-one fashion \parencite{kuznecovElementsAppliedBifurcation2023}. 
Systems that are smoothly orbitally equivalent are also topologically equivalent and share the same number of and type of limit sets (e.g., fixed points, limit cycles). In other words, they have qualitatively the same dynamics and hence implement the same mechanism. 
However, for all but the simplest of systems, finding a diffeomorphism to verify smooth orbital equivalence is highly nontrivial.
As a result, this notion, while fundamentally rigorous and sound, has not been operationalized in practice.

\begin{figure}[htbp]

\begin{center}
\includegraphics[width=\textwidth]{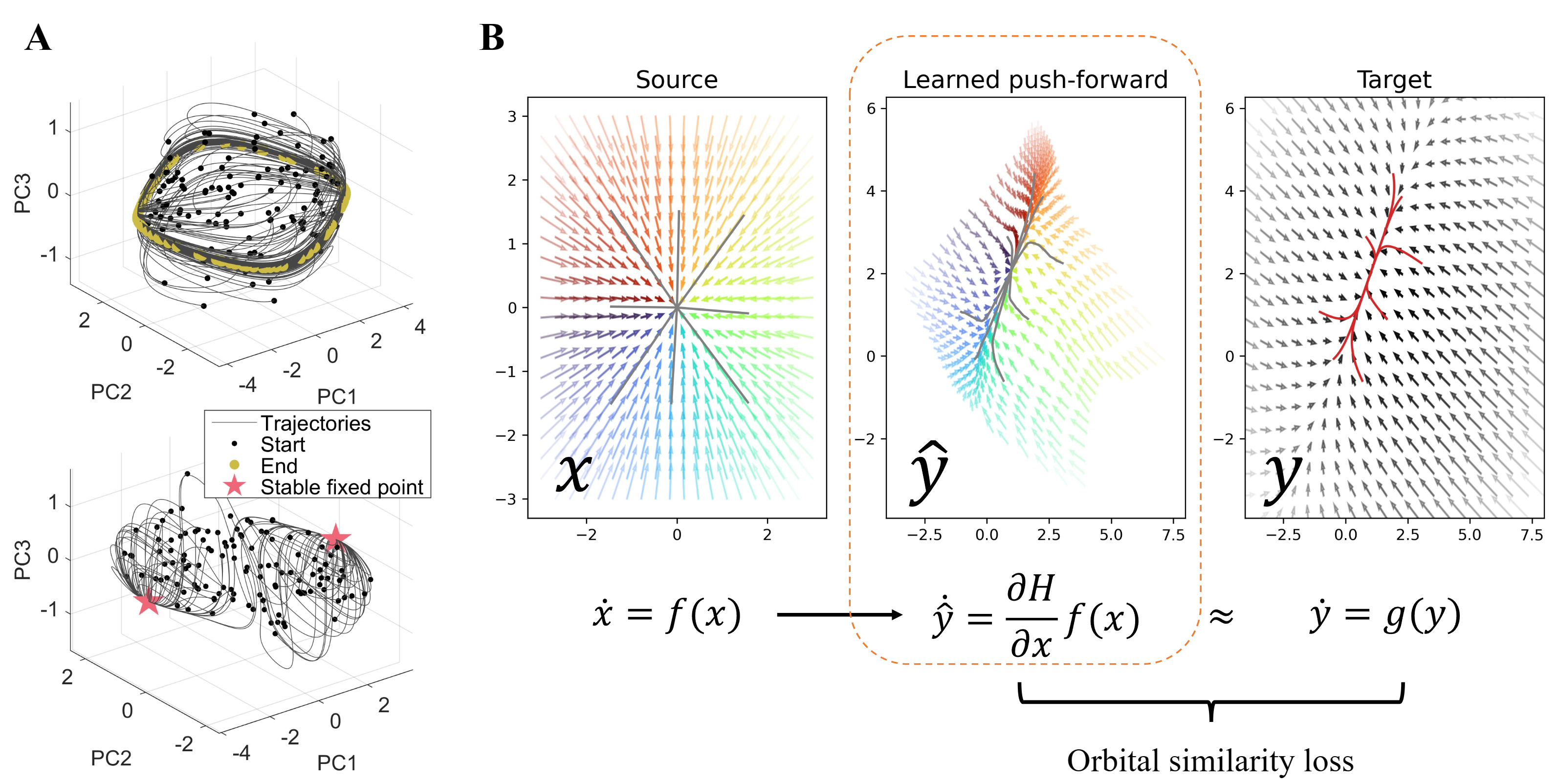}
\end{center}
\caption{{\bf DFORM Schematic.} A: Many efforts to compare learned models resort to assessment of limit sets, often by visualization. Here we plot the projection of simulated trajectories and numerically identified attractors of two models on the first three principal components (PCs) of their trajectories. B: We propose DFORM to learn a diffeomorphism that directly aligns vector fields. The mismatch between transformed and target vector fields defines the orbital similarity loss, which is used to optimize the diffeomorphism.}
\label{fig:schematic}
\end{figure}

Here, we directly address the problem of vector field alignment by learning an orbit-matching diffeomorphism between two systems. To this end, we derive a first-principle loss function using the concept of a \emph{pushforward vector field} (Figure~\ref{fig:schematic}B). We propose modeling a diffeomorphism as the combination of a linear affine mapping and a nonlinear Neural Ordinary Differential Equation \parencite[Neural ODEs,][]{chenNeuralOrdinaryDifferential2018}, which can be effectively optimized through backpropagation using a two-stage training scheme.
We verify the efficacy of our approach on systems related by either linear or nonlinear coordinate transformations.
We then demonstrate that for topologically distinct systems, the end result of DFORM generalizes the notion of orbital equivalence into a continuous index that characterizes how well one system's vector field geometrically maps to that of the other.
Furthermore, we illustrate how DFORM can locate important dynamical motifs on low-dimensional manifolds such as saddle limit sets embedded in high-dimensional models.
Finally, we apply DFORM to a set of models trained to approximate empirical neural data, and showed that DFORM could identify the limit cycles in the models.

The organization of the rest of this paper is the following: First, we review the scientifically and methodologically relevant work and clarify the distinction between DFORM and previous studies. Then, we detail the mathematical theory and computational implementation of DFORM, and present \textit{in silico} experiments of vector field alignment. We proceed to explain how the same framework can be used for identification of dynamical motifs embedded on low-dimensional manifolds, and apply it to various canonical and empirical models. Finally, we summarize our findings and propose several future research directions in the Conclusion and Discussion section.

\section{Related work}

\subsection{Alignment and comparison of dynamical systems}

A common way to characterize and compare high-dimensional dynamical systems is to numerically integrate the trajectories and assess their limit sets. Most frequently, trajectories are visualized after dimensionality reduction to reveal the attractors, which are then used as the criterion for comparison (Figure~\ref{fig:schematic}A).
While often insightful, such analysis is only qualitative and depends solely on the \emph{probable} limit sets (e.g., stable equilibria).
To improve this, \parencite{maheswaranathan2019universality} suggested a quantitative similarity measure based on a transition probability matrix obtained by perturbing around unstable equilibria \parencite[also see][]{smith2021reverse}. The success of such a method relies on the ability to locate unstable equilibria, which, as we will discuss in the next section, has been challenging and might also benefit from DFORM.
Going beyond limit sets, the MARBLE framework \parencite{gosztolai2023interpretable} proposes a similarity measure based on local vector field features sampled from across the phase space. The features are embedded into lower-dimensional space through contrastive learning, then compared using an optimal transport distance.

For models driven by inputs, another way to make comparison is through representation similarity analysis (RSA). Here, `representation' refers to the response of each model to a common input \parencite{kornblithSimilarityNeuralNetwork2019}. Various methods have been proposed to align and compare representations across models, such as Centered Kernel Alignment \parencite[CKA,][]{kornblithSimilarityNeuralNetwork2019} and singular vectors canonical correlations analysis \parencite[SVCCA,][]{raghu2017svcca}.
Like other correlation analyses, RSA assumes independence between different samples (states), as would be the case in feedforward networks where RSA was initially developed. However, dependence between states is the definitive feature of dynamical systems, i.e., how past states lead to the future. Nonetheless, even with the independence assumption, RSA might still capture temporal dependence implicitly, and it can reveal differences between dynamics by aligning and comparing time-locked trial trajectories \parencite{lipshutzDisentanglingRecurrentNeural2024}.
Our method (DFORM) has a fundamentally different philosophy and mathematical formulation relative to RSA.
DFORM operates directly on the \emph{generative} vector fields underlying a dynamical system, rather than on \emph{generated} trajectories driven by inputs. In fact, DFORM does not require simulating any trajectories, and unlike RSA, DFORM can be applied to systems without inputs -(i.e., without a `representation'  \textit{per se}), such as models of resting state brain dynamics \parencite{singhEstimationValidationIndividualized2020, raziLargescaleDCMsRestingstate2017, sipCharacterizationRegionalDifferences2023}.

To the best of our knowledge, the method most directly related to the current paper is the Dynamical Similarity Analysis (DSA) framework \parencite{ostrowGeometryComparingTemporal2023}. DSA learns a linear coordinate transformation which maximizes the cosine similarity between the system matrices of two linear time-invariant systems. For nonlinear dynamical systems, such linear systems can be obtained through Dynamical Mode Decomposition (DMD), which approximates its Koopman eigenspectrum \parencite{schmid2022dynamic}. Importantly, topologically conjugate systems would have the same Koopman eigenspectrum. Assuming that the DMD-approximated linear systems well-characterizes the Koopman eigenspectrum, DSA quantifies how far the two systems are from being conjugate. Despite shared motivation, our work has important conceptual and technical distinctions from DSA as well.
We work directly in the state space of the original nonlinear systems. This means we do not garner any approximation error from DMD or similar linearizing basis expansions.
Critically, we also provide an analytic approximation of the actual diffeomorphism between the two systems. In DSA such a diffeomorphism can only be computed when the DMD mapping can be inverted, which is generally impossible \parencite{bruntonKoopmanInvariantSubspaces2016, bolltMatchingEvenRectifying2018}.

{Finally, a recently developed method called Dynamical Archetype Analysis \parencite[DAA,][]{sagodiDynamicalArchetypeAnalysis2025} also bears some conceptual and methodological similarity with DFORM. DAA learns a nonlinear diffeomorphism between two systems to align and compare them just like DFORM. However, similar to RSA and unlike DFORM, the DAA loss function was based on the mismatch between temporally-aligned \emph{trajectories} rather than the vector fields themselves. Among other distinctions, DAA aims to match both the orientation and the speed of the dynamics, while DFORM is not sensitive to speed differences by default (but can be made so if that is desired; see Discussion). This allows DFORM to align topologically equivalent but non-diffeomorphic systems, for which we will provide many examples throughout the paper.}

\subsection{Identification of dynamical motifs on low-dimensional manifolds}

An important sub-problem that DFORM seeks to engage is the identification of low-dimensional, invariant limit sets, within the high-dimensional system dynamics.
While forward simulation of trajectories can surface asymptotically stable limit sets in state space, identifying the improbable limit sets (e.g., saddle nodes and saddle limit cycles) remains challenging. Theoretically, saddle nodes can be found through traditional root-finding algorithms, but such algorithms usually do not scale well for high-dimensional dynamical models. \parencite{sussillo2014neural} proposed searching for the root of the norm of the ODE (instead of the root of the ODE directly). \parencite{ katzUsingDirectionalFibers2018} suggested another approach by tracing the pre-image of a constant derivative direction (termed \emph{directional fiber}), and found a relatively different set of equilibria than what \parencite{sussilloOpeningBlackBox2013} found. Of course, locating \emph{all} equilibria of a general high-dimensional system remains challenging.
Furthermore, unlike saddle nodes, it is even harder to identify saddle limit cycles, as it cannot be reduced to a root-finding problem without knowing the period of the limit cycle in advance.

DFORM takes a fundamentally different approach to identifying limit sets, relative to root-finding-based methods. Instead of searching for a particular feature in the original high-dimensional state space, DFORM projects the vector field of the original system to a lower-dimensional `template', where the limit sets can be located numerically or even analytically. These limit sets can then be mapped back to the original system through the inverse transformation. Such an approach is particularly suitable for identification of limit cycles with unknown period, since two limit cycles do not need to have the same period for their respective systems to be smoothly orbitally equivalent. Also, since DFORM is solely based on vector field geometry instead of numerical integration, it does not need to assume any stability property of the feature of interest, enabling the identification of improbable features including saddle limit cycles.

A recent, related work is \parencite{friedmanCharacterizingNonlinearDynamics2025}, which proposes a method called \textit{smooth prototype equivalence} (SPE). This methods is similar to the template matching method developed here, and is based on an earlier version of DFORM \parencite{chenDFORMDiffeomorphicVector2024}. 
The template matching capability we show here is more general than SPEs in the sense that we are able to identify limit sets without any assumptions regarding asymptotic convergence. Among other features, this allows for the identification of improbable limit sets, such as saddles.

\subsection{Modeling and optimization of diffeomorphisms}

Constructing a diffeomorphism to minimize the discrepancy between two \emph{scalar} fields has been studied extensively, particularly in image registration problems \parencite{begComputingLargeDeformation2005} and normalizing-flow-based generative modeling \parencite{kobyzevNormalizingFlowsIntroduction2021}.
The emergence of deep learning provides a powerful way to model and optimize more complicated nonlinear diffeomorphisms. While there are many ways to model a diffeomorphism, two popular choices are residual networks \parencite[ResNet,][]{heDeepResidualLearning2015} and Neural Ordinary Differential Equations \parencite[Neural ODEs,][]{chenNeuralOrdinaryDifferential2018}. It is known that the layers of a ResNet can be considered as Euler-discretization of a flow \parencite{rousseauResidualNetworksFlows2020, marionImplicitRegularizationDeep2023}. The invertibility of the ResNet mapping can be enforced as a regularization term \parencite{rousseauResidualNetworksFlows2020}, or by constraining the Lipschitz constant of the residual blocks \parencite{behrmannInvertibleResidualNetworks2019, goukRegularisationNeuralNetworks2021}. On the other hand, Neural ODE models a diffeomorphism as the flow of a time-varying vector field, which is parameterized by a neural network. Such way of constructing diffeomorphisms has a mathematical foundation in shape analysis theory \parencite{younes2019shapes}. Neural ODE significantly simplifies the computation of the inverse diffeomorphism (as the time-reversed flow) and its gradient with respect to the parameters, which is useful for optimizing our loss function. Therefore, we adopted this modeling framework in our study.

Deep-learning-based diffeomorphic alignment has been successfully applied to one-dimensional \parencite[time warping,][]{huangResidualNetworksFlows2021}, two-dimensional \parencite[images,][]{amorResNetLDDMMAdvancingLDDMM2023}, and high-dimensional \parencite[point clouds,][]{battikhKNNResResidualNeural2023} \emph{scalar} fields. However, to our knowledge, it has never been used on \emph{vector} fields. Importantly, the misalignment loss function for scalar and vector fields are fundamentally different. In the former case, the loss only involves the determinant of the Jacobian of the diffeomorphism \parencite[namely the model, see][]{lipmanFlowMatchingGenerative2022}, which can sometimes be approximated with much simpler formula \parencite{behrmannInvertibleResidualNetworks2019}. However, as will be shown below, a proper mismatch loss function for vector fields contains the product of the Jacobian of the model and the vector field under comparison. Whether such loss function can be effectively minimized end-to-end has never been demonstrated before and constitutes a methodological innovation in the current work.

\section{Diffeomorphic vector field alignment}

In this section, we lay out the mathematical and computational basis of DFORM. We then apply DFORM to align and compare dynamical models using linear transformations. Finally, we demonstrate DFORM's ability to {nonlinearly align and compare} two systems.

\subsection{Vector field alignment as an optimization problem} \label{sec:DFORM}

We consider the problem of aligning and comparing learned dynamical models. We denote any two such models as $\dot x = f(x)$ and $\dot y = g(y)$, where $x,y \in\mathbb{R}^{n}$. Note that exogenous inputs/stimuli that are constant or piece-wise constant (as is common in theoretical neuroscience) are readily absorbed into the formulation of $f$ and $g$ (see also Discussion).
From a dynamical systems perspective, these models are said to be \emph{smoothly orbitally equivalent}, should there exist a (smooth) diffeomorphism (a smooth map with smooth inverse) $\varphi: \mathbb{R}^n \to \mathbb{R}^n$ such that
\begin{equation}\label{eq:equivalency}
    \frac{\partial \varphi}{\partial x}f(x) = c(x)g(\varphi(x))\,,
\end{equation}
where $c(x) > 0$ is a smooth function. If $c(x) \equiv 1$, the two systems are called \emph{diffeomorphic} \parencite{kuznecovElementsAppliedBifurcation2023}. {Note that the Jacobian matrix $\left[\frac{\partial \varphi}{\partial x}\right]_{ij} := \frac{\partial \varphi_{i}}{\partial x_j}$ and the vector field $f$ are always evaluated at the same point (here $x$).}

This can be understood geometrically if we define the \emph{pushforward} (`coordinate-transformed') vector field of $f$ by $\varphi$ as:
\begin{equation}
    \varphi_{*}f(y) := \frac{\partial \varphi}{\partial x}f(\varphi^{-1}(y))\,,
\end{equation}
where $y = \varphi(x)$. Eq.(\ref{eq:equivalency}) can then be written as $\varphi_{*}f(y) = c_{y}(y)g(y)$. Therefore, $f$ and $g$ are smoothly orbitally equivalent if there exists a coordinate transformation $\varphi$ such that the pushforward $\varphi_{*}f$ and the target $g$ differ only in magnitude (and smoothly so) but not in direction (Figure~\ref{fig:schematic}B). If $\varphi_{*}f = g$ instead, then $f$ and $g$ are diffeomorphic.

In practice, the evaluation of $\varphi$ is intractable for all but the simplest of dynamical systems.
In order to operationalize this comparison, we propose DFORM, in which a neural network is trained to learn $\varphi$. We introduce the \emph{orbital similarity loss}:
\begin{equation}\label{eq:topoloss}
\begin{split}
    J_{f, g, \varphi}(y) & = \frac{1}{n}\left\| \frac{\frac{\partial \varphi}{\partial x}f\big(\varphi^{-1}(y)\big)}{\left\| \frac{\partial \varphi}{\partial x}f\big(\varphi^{-1}(y)\big) \right\|_{2}} - \frac{g(y)}{\|g(y)\|_{2}} \right\|_2^2 \\
    & = \frac{1}{n}\left\| \frac{\varphi_{*}f(y)}{\left\|\varphi_{*}f(y)\right\|_2} - \frac{g(y)}{\|g(y)\|_{2}} \right\|_2^2\,.
\end{split}
\end{equation}
This loss is minimized when $\varphi_{*}f$ and $g$ differ in magnitude but not direction. In this case, $\varphi$ exactly maps the $x$ orbits to the $y$ orbits in a one-to-one fashion.

One can also define a loss function in the state space of $f$ instead of that of $g$:
\begin{equation}\label{eq:topoloss_inv}
    J_{g, f, \varphi^{-1}}(x) = \frac{1}{n}\left\| \frac{(\varphi^{-1})_{*}g(x)}{\left\| (\varphi^{-1})_{*}g(x) \right\|_2} - \frac{f(x)}{\|f(x)\|_{2}} \right\|_2^2\,,
\end{equation}
which characterizes the mismatch between $f$ and the inverse pushforward function $(\varphi^{-1})_{*}g$. It is worth noting that the two terms Eq.(\ref{eq:topoloss}) and Eq.(\ref{eq:topoloss_inv}) are usually numerically different (see Appendix~\ref{app:symmetry}).
To provide an interpretable alignment when $f$ and $g$ are not necessarily equivalent, one would like to make sure that both terms are small. Therefore, we minimize the sum of the two terms
\begin{subequations} \label{eq:master_loss}
\begin{align}
    l_{1} &= \textbf{E}_{y}\left[ J_{f, g, \varphi}(y) \right]\,, \label{eq:master_loss_1} \\
    l_{2} &= \textbf{E}_{x}\left[ J_{g, f, \varphi^{-1}}(x) \right]\,, \label{eq:master_loss_2}
\end{align}
\end{subequations}
where $\mathbf{E}_{x}$ and $\mathbf{E}_{y}$ indicate the expected values over $x\sim p_{x}$ and $y\sim p_{y}$ respectively, for some distribution $p_{x}$ and $p_{y}$ (see {Discussions and} Appendix~\ref{app:sampling}).

{
Note that the orbital similarity loss Eq.(\ref{eq:topoloss}) naturally relates to the \emph{cosine similarity} between the pushforward vector field $\varphi_{*}f$ and the target $g$. Denote the directional vectors $\mathbf{v}:= \frac{\varphi_{*}f(y)}{\left\|\varphi_{*}f(y)\right\|_2}$ and $\mathbf{w} := \frac{g(y)}{\|g(y)\|_{2}}$, and let $\angle(\mathbf{v}, \mathbf{w})$ indicate the angle between $\mathbf{v}$ and $\mathbf{w}$. Then we have $J_{f, g, \varphi}(y) = \frac{1}{n}\| \mathbf{v} - \mathbf{w} \|_2^2 = \frac{2}{n}(1 - \mathbf{v}^{\top} \mathbf{w}) = \frac{2}{n}\left(1 - \cos\angle\left(\varphi_{*}f(y), g(y)\right)\right)$.}

Therefore, the extent of alignment achievable through training $\varphi$ provides a natural generalization of the notion of orbital equivalence.
{
After training, we define the \emph{forward alignment} between $\varphi_{*}f$ and $g$ as
\begin{equation} \label{eq:forward_alignment}
    \textbf{E}_{y}\left[\cos\angle\left(\varphi_{*}f(y), g(y)\right)\right]\,,
\end{equation}
and the \emph{backward alignment} between $f$ and $(\varphi^{-1})_{*}g$ as
\begin{equation} \label{eq:backward_alignment}
    \textbf{E}_{x}\left[\cos\angle\left(f(x), (\varphi^{-1})_{*}g(x)\right)\right]\,.
\end{equation}
And we define the \emph{orbital similarity} between $f$ and $g$ as the minimum of the two:
\begin{equation}\label{eq:similarity}
    \min\left\{\textbf{E}_{x}\left[\cos\angle\left(f(x), (\varphi^{-1})_{*}g(x)\right)\right], \textbf{E}_{y}\left[\cos\angle\left(\varphi_{*}f(y), g(y)\right)\right]\right\}\,.
\end{equation}
}

\subsection{Implementation with Neural Ordinary Differential Equations} \label{sec:NeuralODE}

To model and optimize the diffeomorphism $\varphi$, we note that the set of all smooth diffeomorphisms over $\mathbb{R}^{n}$ is an infinite-dimensional Lie group $\text{Diff}(\mathbb{R}^{n})$ with the exponential mapping being the flow of vector fields \parencite{kriegl1997convenient}. Therefore, we can naturally construct a diffeomorphism as the flow of a (time-varying) vector field $\varphi(x) = \Phi_{01}^{v}(x)$, where $x(t) = \Phi_{0t}^{v}(x^{0})$ is the solution of the ODE $\dot{x} = v(t, x), x(0) = x^{0}$ \parencite{younes2019shapes}.
Such diffeomorphisms will always be orientation-preserving and cannot model reflections such as $y = -x$ for $x \in \mathbb{R}^{2k + 1}, k\in \mathbb{N}$. To be more comprehensive, here we represent a diffeomorphism as the composition of an invertible linear affine transformation and the flow of a nonlinear vector field:
\begin{equation}
    \varphi = \Phi_{01}^{v}\circ\mathcal{H}\,,
\end{equation}
where $\mathcal{H}(x) = Hx + b, \det H \ne 0$. Conceptually, $\varphi$ first translates, reflects, scales and rotates the phase space through a linear affine transformation, then stretches the phase space nonlinearly along the trajectories of the vector field $v$.

It can be shown that the solution $\varphi$ to the alignment problem is not unique (see Appendix~\ref{app:identifiability}). To reduce redundancy and facilitate convergence, we set $v$ to be time-invariant throughout the rest of this paper. We regularize the norm of $v$ to push the nonlinear transformation towards identity mapping \parencite{salmanDeepDiffeomorphicNormalizing2018}, and we also regularize the Frobenius norm of $HH^{T} - I_{n}$ to push the linear transformation towards orthogonal \parencite{brockNeuralPhotoEditing2017}. Due to regularization, $\varphi$ can be understood as a small nonlinear deformation on top of a linear affine transformation, delineating the contribution of nonlinear alignment over and above the linear one {(see Appendix~\ref{app:sampling})}.

We implemented the method using the Neural Ordinary Differential Equation (Neural ODE) framework \parencite{chenNeuralOrdinaryDifferential2018}. A two-hidden-layer feedforward neural network learns the deformation vector field $v(x)$. The width of the input and output layer is the dimensionality of the systems, $n$. The width of each hidden layer is set to $2n$ or 20, whichever is larger. We used the exponential linear unit (ELU) activation function \parencite{clevertFastAccurateDeep2016}, which guarantees the smoothness of the transformation. The nonlinear flow is numerically integrated using Runge-Kutta method of order 5 of Dormand-Prince-Shampine, with a relative tolerance of $10^{-5}$. The linear transformation $\mathcal{H}(x) = Hx + b$ is learned by a linear layer with $n$ inputs and $n$ outputs. The total number of parameters is thus $n^{2} + 42n + 440$ for $n < 10$ and $9n^{2} + 6n$ for $n \ge 10$.

{The model were trained in (mini-)batches. In each batch, a certain number of random samples were drawn from the phase space of $f$ and $g$ respectively to evaluate the expected value of Eq.(\ref{eq:master_loss_1}) and Eq.(\ref{eq:master_loss_2}) (see Appendix~\ref{app:sampling}). 
The inverse function $\varphi^{-1} = \mathcal{H}^{-1}\circ\Phi_{10}^{v}$ was computed by integrating the flow backward in time. Jacobian-vector products were computed through forward-mode auto-differentiation in \texttt{PyTorch}.}
The training process consisted of two phases. In the first phase, we fixed the nonlinear transformation to be identity (namely $v(x)\equiv \mathbf{0}$) and trained the linear transformation $\mathcal{H}$ for certain number of batches. The model weights were updated with a NADAM optimizer with a default learning rate of 0.002 \parencite{dozatIncorporatingNesterovMomentum2016}.
This pre-training provided a good linear approximation of the nonlinear solution that facilitated the convergence of the full-scale training. In the second phase, we trained all the weights for another certain number of batches with another NADAM optimizer with a smaller learning rate of $0.0002$.
The largest-scale experiment we tried involved two 100-dimensional systems with 2000 batches of full scale training and took around 20 minutes to finish on a Nvidia V100 GPU. The hyperparameter setting for each experiment was summarized in Appendix~\ref{app:hyperparameter}.

\subsection{Alignment and comparison of vector fields through linear transformation}

In the following sections, we conduct experiments to align and compare dynamical models using DFORM. While we designed DFORM as a general tool for learning nonlinear coordinate transformations between vector fields, it is useful to begin with linear transformations as it allows us to avoid the subtleties of the optimization process and focus on understanding the nature of the vector field alignment problem itself. In this section, we present experiments where we used a linear DFORM model to align topologically equivalent systems or to compare topologically non-equivalent systems. Results demonstrated several surprising properties of the vector field alignment problem, such as the existence of multiple optimal solutions, and the influence of sample distributions.

\subsubsection{{Linear orbital alignment between topologically equivalent linear systems}} \label{sec:linear_equiv}

Denote two linear systems as $\dot{x} = f(x) = A_{1}x$ and $\dot{y} = g(y) = A_{2}y$. If a matrix $A$ has $p$ eigenvalues with positive real parts, $q$ with negative real parts and $r$ with zero real parts, we say that $A$ has a \emph{signature} of $(p, q, r)$. It is a standard result in dynamical system theory that two linear systems are topologically equivalent if and only if $A_{1}$ and $A_{2}$ have the same signature. However, the homeomorphism between them is usually nonlinear and not differentiable both ways (thus not a diffeomorphism). In fact, two linear systems are diffeomorphic if and only if they are \emph{linearly equivalent}, i.e., there exists an invertible matrix $H\in\mathbb{R}^{n\times n}$ such that $A_{2} = HA_{1}H^{-1}$. In this case, the coordinate transformation will be given by $y = Hx$.

To test how well one can linearly align different kinds of topologically equivalent linear systems, we consider four cases based on the relationship between the two systems' eigenspectrum: (1) $A_{2} = HA_{1}H^{-1}$ and $H$ is orthogonal; (2) $A_{2} = HA_{1}H^{-1}$ and $H$ is invertible but not necessarily orthogonal; (3) $A_{1}$ and $A_{2}$ have the same number of positive/negative/zero eigenvalues, as well as the same number of complex eigenvalues with positive/negative/zero real parts (referred to as `same type' for simplicity below); (4) $A_{1}$ and $A_{2}$ have the same signature but not necessarily the same type of eigenvalues (referred to as `same sign', e.g., $A_{1}$ has two positive real eigenvalue and $A_{2}$ has a pair of complex eigenvalue with positive real parts). The first two cases require a linear transformation and the last two require a nonlinear one. As we try to align the systems using a linear transformation, it is expected that the first two cases should result in very good alignment and the last two cases less so. The differences between them will indicate the limitation of linear methods in vector field alignment.

We randomly generated 30 pairs of systems for each of the four categories (Orthogonal, Linear, Same type, and Same sign; see Appendix~\ref{app:linear-systems} for details), and trained a linear DFORM (i.e., only including the linear layer) between each pair of systems. {Training was repeated with three different initializations and the solution with the highest orbital similarity was retained.} In each batch, we drew 128 samples of $x$ and 128 samples of $y$ from the standard normal distribution with identity covariance to calculate the orbital similarity loss. We trained DFORM for 2500 batches until convergence. {We quantified the forward alignment} between $\varphi_{*}f$ and $g$ {as in Eq.(\ref{eq:forward_alignment})}, where $\varphi$ represents the DFORM transformation. We had also tried to calculate the cosine similarity between the Jacobian matrices of $\varphi_{*}f$ and $g$ at the origin and obtained almost identical results, indicating consistency between our method and Jacobian-alignment-based methods \parencite[e.g., DSA;][]{ostrowGeometryComparingTemporal2023} on linear systems.

\begin{figure}[htbp]

\begin{center}
\includegraphics[width=0.8\textwidth]{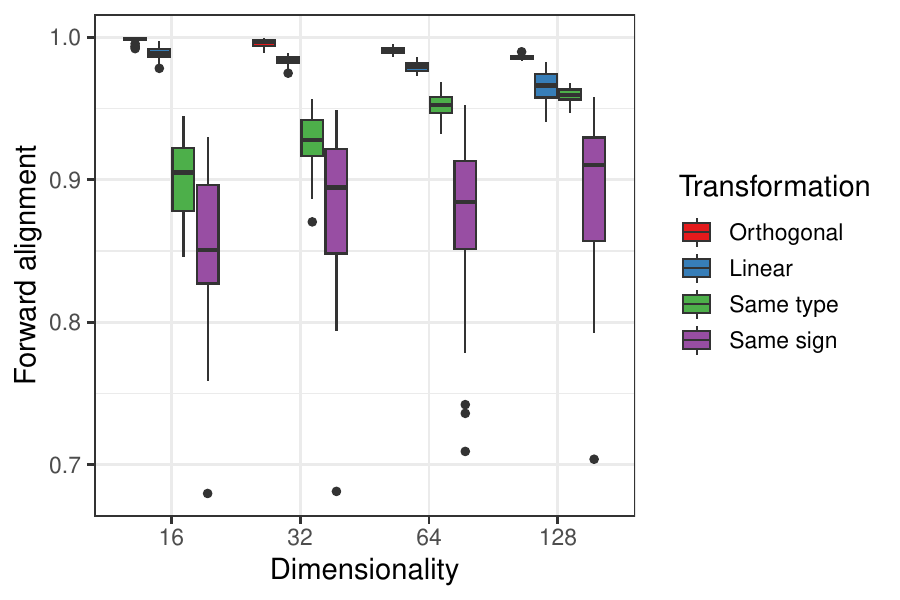}
\end{center}
\caption{{\bf Identification of transformations between equivalent linear systems.} We randomly generated pairs of topologically equivalent linear systems of different sizes. Box plots showed the distribution of {forward alignment} between $\varphi_{*}f$ and $g$ after DFORM training, across 30 experiments for each of the four types of linear systems described in the main text (indicated by different colors). Bottom, middle and top horizontal lines in each plot showed 25, 50, and 75 percentiles respectively. Whiskers extend to the largest/smallest value within 1.5 times of interquartile range from the hinges. Dots represent outliers outside this range.}  %
\label{fig:linear-equiv}
\end{figure}

Results were summarized in Figure~\ref{fig:linear-equiv}. As expected, the {forward alignment} between $\varphi_{*}f$ and $g$ was higher for linearly-related systems (orthogonal or general linear), which remained above {0.95} even at $n = 128$. The {alignment} for nonlinearly-related systems was not as high but still mostly above 0.8, suggesting that linear alignment could also provide a decent approximation of the actual nonlinear transformation between equivalent systems. As we will show in the next section, {an alignment} of 0.8 is higher than what we usually can obtain between non-equivalent systems. Therefore, linear vector field alignment can be used to infer the diffeomorphic and topological equivalency between linear systems.

Perhaps surprisingly, while DFORM achieved good alignment between linearly transformed systems (orthogonal or non-orthogonal), the transformations it found were in general very different from the ground-truth one (see Supplementary Figure~\ref{fig:linear-scatter}). This result revealed an important property of the vector field alignment problem, i.e., the existence of multiple optimal solutions. In fact, we can prove that for a linear system $f(x) = A_{1}x$ and a given linear transformation $\mathcal{H}_{1}(x) = H_{1}x$, we can take an arbitrary invertible polynomial function $P(A_{1})$ of $A_{1}$ and define a new linear transformation $\mathcal{H}_{2}(x) = H_{1}P(A_{1})x$, such that the pushforward vector field $(\mathcal{H}_{1})_{*}f$ and $(\mathcal{H}_{2})_{*}f$ are identical (see Appendix~\ref{app:identifiability} for formal analysis). Theoretically, this property suggests that it is possible to identify the ground truth transformation between linear systems only up to an equivalence class. As we show in the following sections, this property has important implications for nonlinear systems alignment, where linearization-based methods may fail due the multiplicity of solutions.

\subsubsection{{Linear orbital alignment between topologically non-equivalent linear systems}} \label{sec:linear_sgndiff}

As demonstrated by the examples above, the degree of alignment achievable through DFORM was influenced by several factors including the type of transformation, the size of the systems, etc. Therefore, to use DFORM as a tool to suggest the equivalency between two systems (or the lack thereof), it is important to understand the distributions of orbital similarity when the systems under comparison are either equivalent or nonequivalent, respectively.

\begin{figure}[htbp]

\begin{center}
\includegraphics[width=\textwidth]{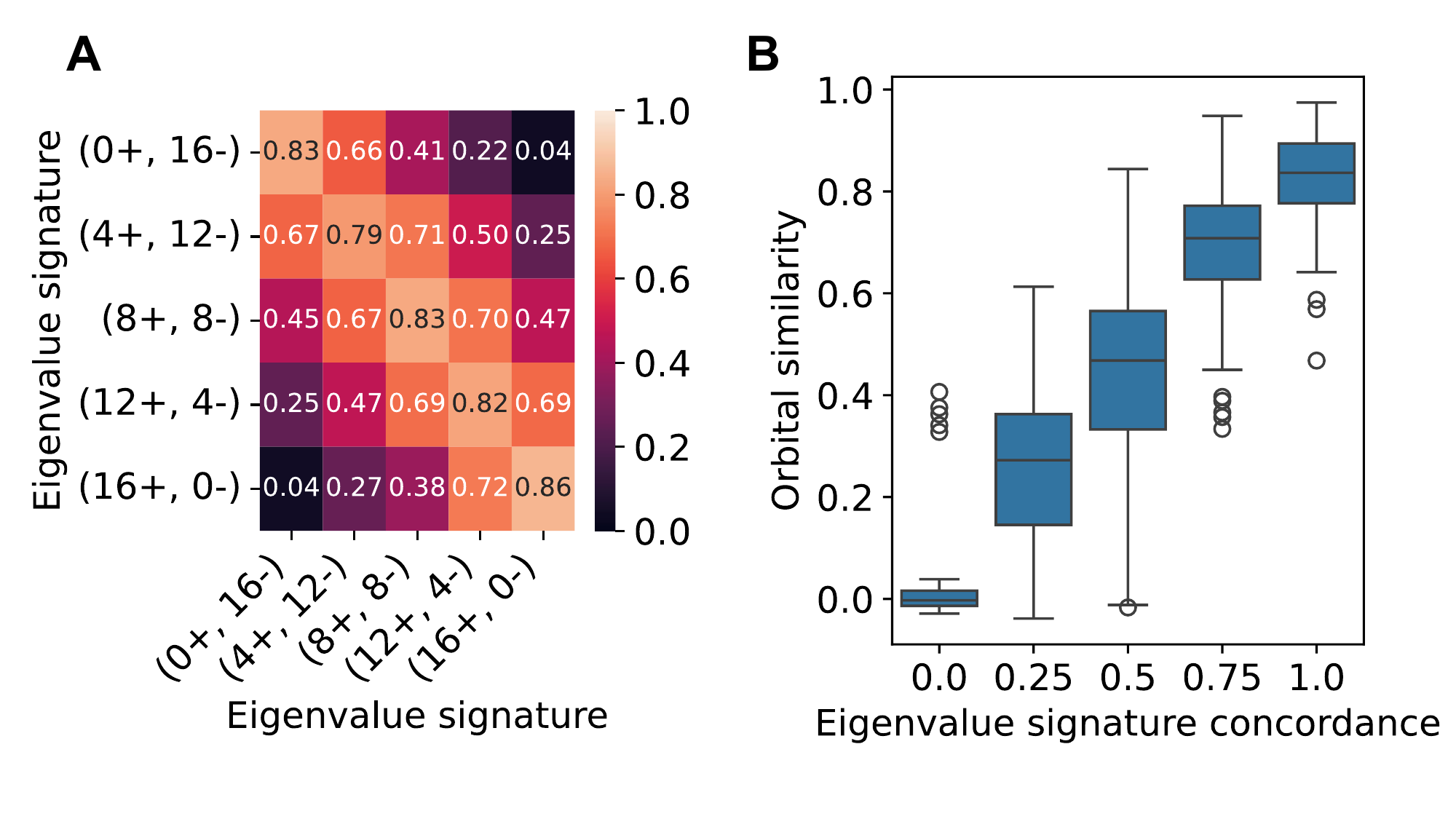}
\end{center}
\caption{{\bf Orbital similarity between linear systems with different signatures.} A. Mean orbital similarity after DFORM alignment between linear systems. X and Y values indicate the signature of the first and second system under comparison, respectively. The signature was represented as an ordered pair, indicating the number of eigenvalues with negative and positive real parts. Color and number in each block indicate the mean orbital similarity across 20 experiments. B. Distribution of the similarity based on the concordance between the two systems' signatures. The five concordance levels are indicated by values along the horizontal axis and correspond to the fourth super/sub-diagonals to the main diagonal in panel A, respectively.}
\label{fig:sgndiff}
\end{figure}

To this end, we constructed many 16-dimensional linear systems and analyzed the orbital similarity between them through linear DFORM training {as in Eq.(\ref{eq:similarity})}. Since two linear systems are topologically equivalent if and only if their system matrices have the same signature, we hypothesized that the orbital similarity between linear systems should reflect the concordance between their signatures. Therefore, we considered the following signatures: $(16, 0, 0)$, $(12, 4, 0)$, $(8, 8, 0)$, $(4, 12, 0)$, and $(0, 16, 0)$. For simplicity, we did not consider systems with zero or purely imaginative eigenvalues, as such cases are statistically rare. We randomly constructed two linear systems with signatures selected from the list above independently (see Appendix~\ref{app:linear-systems} for details), and trained a DFORM between the two systems using the same setting as in the last section. Training was repeated {with three different initializations} and the model with the highest orbital similarity was selected. We conducted 20 experiments for each of the $5^2 = 25$ ordered pairs of signatures and calculated the {orbital similarity}, as shown in Figure~\ref{fig:sgndiff}A. The mean similarity for topologically equivalent systems (i.e., the ones with the same signature) were beyond 0.8 in general, while the similarity between non-equivalent systems gradually decreased as the concordance between their signatures decreased (from the main diagonal to the fourth super/sub-diagonals). To visualize this trend better, we plotted the distribution of orbital similarity across all experiments according to the concordance between the signatures of the systems under comparison. Here we define concordance as $\frac{n - |p_{1} - p_{2}|}{n}$ where $p_{1}$ and $p_{2}$ are the number of eigenvalues with positive real parts for system one and two, respectively, and $n = 16$ is the size of the systems. Therefore, full concordance implies the same signature and zero discordance implies that one system's eigenvalues have all positive real parts while the other's have all negative real parts. As expected, the orbital similarity increased as the two signatures became more concordant. This result suggests that orbital similarity after DFORM mapping can be used to index topological congruency, at least for linear systems.

\subsubsection{{Linear orbital alignment between nonlinear systems}} \label{sec:equiv_RNN}

Going beyond linear systems, in this section we will explore the alignment between nonlinear systems using linear transformations, and proceed to full nonlinear transformations in the next section. We constructed a series of nonlinear systems using RNN models with `random plus low rank' connectivity structure (see Appendix~\ref{app:RNN}). This type of model has been widely used in theoretical neuroscience studies \parencite{mastrogiuseppeLinkingConnectivityDynamics2018}. Under some constraints, such a model can show various kinds of dynamics, including bistability and oscillation \parencite{schuesslerDynamicsRandomRecurrent2020}. We applied random linear transformations (either orthogonal or non-orthogonal) on the models and trained a linear DFORM to align the original system $f$ and transformed systems $g$. Inspired by the findings in the previous sections, we calculated two types of similarity measure to quantify the alignment between DFORM-transformed system $\varphi_{*}f$ and target system $g$: (1) the cosine similarity between the Jacobian matrices at the origin; and (2) the cosine similarity between the location of the nonzero stable fixed points (if any). DFORM was trained for 20,000 batches. In each batch, we drew 32 samples of $x$ and $y$ from the standard normal distribution with identity covariance. Training was repeated five times using different random initializations, and the model with best fixed point alignment was selected.

\begin{figure}[htbp]

\begin{center}
\includegraphics[width=\textwidth]{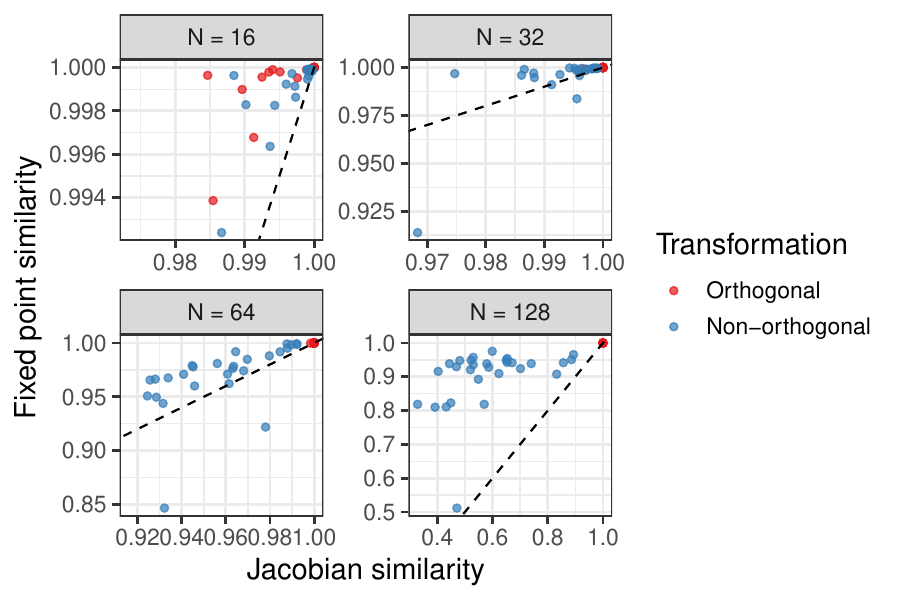}
\end{center}
\caption{{\bf Identification of transformations between nonlinear systems.} We generated RNNs with `low-rank plus random' connectivity and applied a random linear transformation (either orthogonal or non-orthogonal, as indicated by colors) to each system. Scatter plots showed the obtained Jacobian similarity (X axis) and fixed point similarity (Y axis) between each pair of DFORM-transformed and target systems. Experiments were grouped into panels by the size of the systems (16 to 128). Systems without nonzero stable equilibria were not shown in the figure as the fixed point similarity was meaningless, but their Jacobian similarity followed similar distribution as the ones shown.}
\label{fig:RNN}
\end{figure}

Results were shown in Figure~\ref{fig:RNN}. When the ground-truth transformation was orthogonal, DFORM was able to recover it even with system dimension $n = 128$, as indicated by near-perfect cosine similarity between either the fixed points or Jacobian matrices of the transformed and target system. When the transformation is non-orthogonal, {DFORM can also provide decent alignment, with fixed point alignment above 0.9 in general.}

Interestingly, {if we examine the learned DFORM models over all initializations (instead of only the best out of five; Supplementary Figure~\ref{fig:RNN-scatter-all})}, there exist {suboptimal solutions} where the Jacobian similarity was very high but the fixed point similarity was low.
To understand this phenomenon, consider the linearized system $\dot{x} = J_{1}x$ of $f$ at the origin (which is always a fixed point), where $J_{1}$ is the Jacobian of $f$ at the origin. Obviously, the linearized system of $g$ will be $\dot{y} = J_{2}y$ where $J_{2} = H_{0}J_{1}H_{0}^{-1}$, with $H_{0}$ being the ground truth transformation. 
As we discussed in previous sections and shown in Appendix~\ref{app:identifiability}, there exists an infinite number of $H = H_{0}P(J_{1})$ that result in the same transformed Jacobian $J_{2} = HJ_{1}H^{-1} = H_{0}J_{1}H_{0}^{-1}$. When the samples for loss calculation were mostly drawn around the origin, both $f$ and $g$ can be well-approximated by the linearized system. Therefore, DFORM might converge to any of such $H$ as a local minimum. However, such $H$ probably do not align the nonzero fixed points of the systems well, since they are related by $H_{0}$ instead.
{Within the DFORM framework, this issue can be usually resolved by selecting the best model out of several random initializations. As shown in Figure~\ref{fig:RNN}, five repetitions were enough for decent fixed point alignment. However, note that one would not be able to even compute fixed point alignment if a coordinate transformation is not available, e.g., in DSA \parencite{ostrowGeometryComparingTemporal2023}.}

Importantly, because the nonzero fixed points could be completely mismatched even when the Jacobian similarity is high, to correctly align two nonlinear systems, it will not be enough to simply align the linearized systems. We prove that this conclusion is true as long as the linearization is not exact, even when it is based on methods other than Jacobian linearization, such as basis function expansion (see Appendix~\ref{app:linearization}). This observation reveals a crucial limitation in linearization-based method for nonlinear system alignment, and suggests the importance of methods that respect the nonlinear nature of the systems, such as DFORM.

\subsection{{Alignment and comparison through} nonlinear transformations}

In the previous sections, we introduced the vector field alignment problem and demonstrated the effectiveness of the DFORM framework for linear matching. However, we also revealed the fundamental limitation of the linear matching framework. In this section, we present two types of experiments to verify DFORM's ability to identify nonlinear coordinate transformations, one is a canonical example of low-dimensional bifurcating systems, and the other is to {align systems transformed by nonlinear flows. We continue to show that orbital similarity after nonlinear alignment reflects both topological congruency and geometrical similarity, using a bifurcating line attractor model as example.}

\subsubsection{Van der Pol oscillators}  \label{sec:equiv_VDP}

\begin{figure}[htbp]

\begin{center}
\includegraphics[width=\textwidth]{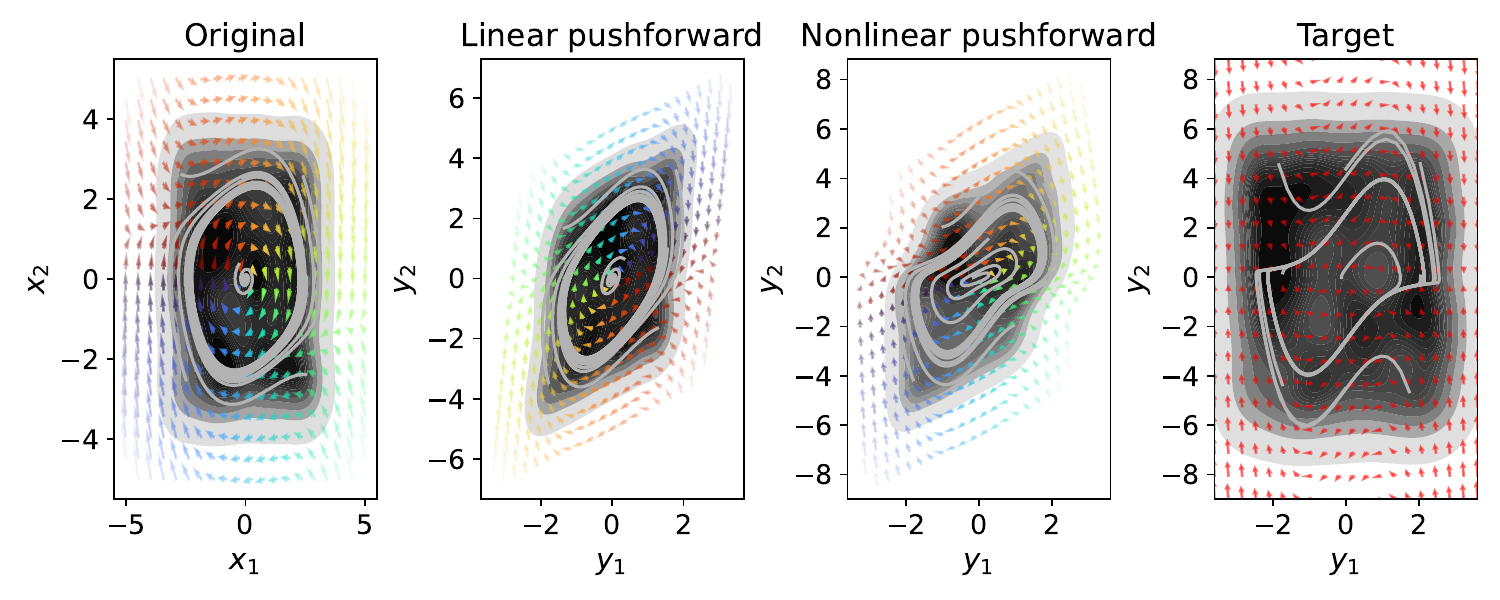}
\end{center}
\caption{{\bf Identification of nonlinear transformation between Van der Pol oscillators.} We generated two Van der Pol oscillator systems (see methods) with different bifurcation parameters $\mu = 0.2$ (left) and $\mu = 2$ (right).  A linear DFORM model (middle left) and a nonlinear DFORM model (middle right) were trained using the same number of samples. Left and right panels: Dark contour plots visualized the sample distributions. Arrows showed the direction and magnitude of the vector fields (length normalized within each panel). Gray curves represented simulated trajectories. Middle panels: Pushforwards of the vector fields, distributions and trajectories in the left panel by the learned DFORM model. Arrows were colored according to their preimages in the left panels.}
\label{fig:VDP}
\end{figure}

The Van der Pol oscillator (Appendix~\ref{app:VDP}) is a widely-used model in many fields. The dynamics of the system are controlled by a bifurcation parameter $\mu$. We generate two Van der Pol oscillator systems with bifurcation parameter $\mu$ being 0.2 and 2 respectively. As shown in Figure~\ref{fig:VDP}, both systems have a globally stable limit cycle, but the shapes of two limit cycles are different in a nonlinear way.

We trained both a linear DFORM model and a nonlinear one between the two systems. The linear model was trained for 5000 batches of 32 samples from each system. The nonlinear model was first pretrained for {2000} batches with the nonlinear part fixed as identity function, then underwent full-scale training for another {3000} batches. Both models had converged by the end of training. For both systems, samples were drawn from a uniform distribution over the rectangle area with $x$-coordinate between $[-3, 3]$ and $y$-coordinate between $[-1.5\mu - 3, 1.5\mu + 3]$, where $\mu$ is the bifurcation parameter. Such sample distribution contains the limit cycle of the system, as shown by the dark contour plot in the left and right panels in Figure~\ref{fig:VDP}. Because both linear and non-linear DFORMs utilized the same number of samples from the same distribution and both converged, the differences in performance should reflect meaningful differences in their capability to approximate nonlinear coordinate transformations, rather than nuances in optimization. While the linear method provided decent alignment (middle left, Figure~\ref{fig:VDP}), the nonlinear method (middle right, Figure~\ref{fig:VDP}) clearly captured the irregular warping of the limit cycle much better, as confirmed by higher {forward alignment} between transformed and target vector fields (linear method: {0.904}; nonlinear method: {0.971}). This result demonstrated that while linear DFORM already provides decent alignment for common examples, nonlinear DFORM is capable of identifying more realistic nonlinear transformations.

\subsubsection{{Alignment of flow-transformed systems}} \label{sec:paramrec}

\begin{figure}[htbp]

\begin{center}
\includegraphics[width=0.85\textwidth]{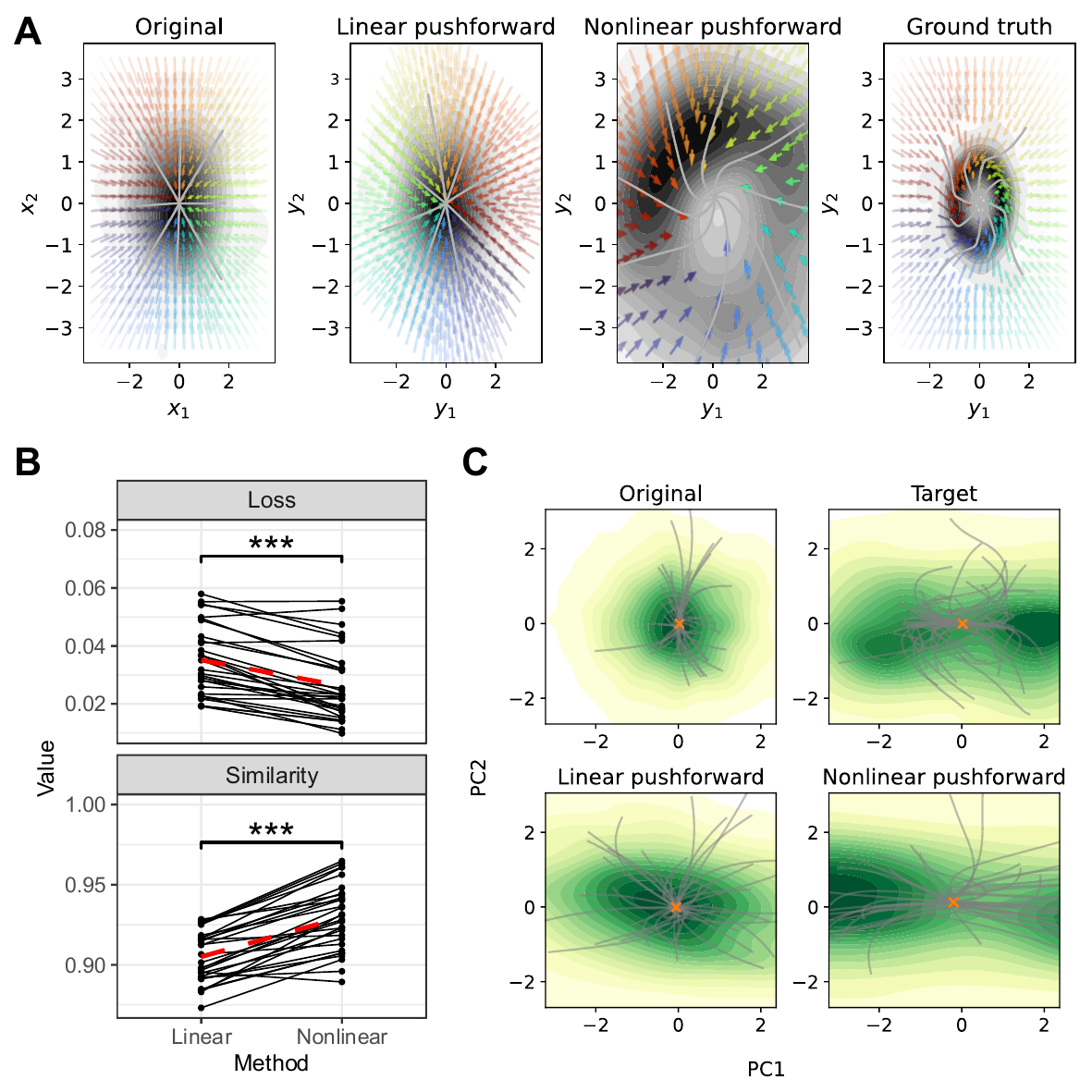}
\end{center}
\caption{{\bf {Learning} nonlinear transformation generated by {flows}.} {A. Alignment of two systems related by a nonlinear flow. Left panel: vector field, simulated trajectories and sample distribution for system $f$. Other panels: the pushforwards by learned linear transformation, nonlinear transformation, and ground-truth transformation respectively (from left to right). Same convention as in Figure~\ref{fig:VDP}.} {B.} Final orbital alignment loss and orbital similarity after DFORM training. Each line represents one experiment. Red dashed lines indicate the mean across 30 experiments. {C.} Comparison of the solution found with linear and nonlinear DFORM. {Top right: 50 trajectories of system $g$ were simulated and projected to the first two principal components (PCs), visualized as gray traces with the end points shown as orange crosses. Green contour plots showed the projection of sample distribution on the PCs. Top left: trajectories and sample distribution of $f$ in the same PC space, with initial conditions being the inverse transformation of those of $g$'s. Bottom: the pushforward of the trajectories and distribution of $f$ by the learned transformations.}}
\label{fig:param-rec}
\end{figure}

{In this section, we provide two examples where the ground-truth transformation $\varphi_{0}$ is given by the flow $\Phi_{01}^{v_{0}}$ of a ground-truth vector field $v_{0}$. In the first experiment, we define $v_{0}$ to be the canonical form of supercritical Hopf bifurcation with parameter $\mu = 1$ (see Appendix~\ref{app:Hopf}). This system contains a globally stable limit cycle at the unit circle. To handle numerical issues during integration, we further multiplied $v_{0}$ with a damping factor $e(x) = \exp(\frac{-||x||_{2}}{2})x$, so that the deformation shrank to zero at the infinity. We generated a stable linear system $f$ and pushed it forward by $\varphi_{0}$ to obtained system $g$. As visualized in Figure~\ref{fig:param-rec}A, $\varphi_{0}$ `stretches' the trajectories in a spiral towards the unit circle from both the inside and the outside, making it fundamentally nonlinear.}

{We then trained both a linear and a nonlinear DFORM to learn a transformation between the two systems. To be more straightforward, we removed the linear layer from the nonlinear DFORM, and trained both models for 2000 batches with an initial learning rate of 0.001. In each batch, 32 random $x$ were drawn from the standard normal distribution with identity covariance, and 32 random $y$ were generated by $y = \varphi_{0}(x)$ with another 32 random $x$ drawn from the same distribution. The learned pushforwards were visualized in Figure~\ref{fig:param-rec}A. The nonlinear DFORM clearly matched the vector field and trajectories better, and it successfully captured the transport of probability mass away from the origin. Interestingly, while the orbital similarity was very high (above 0.95), the learned pushforward distribution did not match the ground truth. This finding is, once again, explained by the multiplicity of solutions to the vector field alignment problem. In fact, transforming a stable linear system $f$ by $\varphi$ results in the same pushforward vector field as transforming by $\varphi\circ\Phi_{0\tau}^f$, yet the pushforward distribution will be more dispersed if $\tau < 0$, similar to what we found here.}

Apart from {this illustration, here we also} provide a more general and higher-dimensional example. We considered an eight-dimensional random linear system $f$ and transformed it to $g = (\varphi_{0})_{*}f$ by a nonlinear random DFORM network $\varphi_{0}$, {parameterized by} a two-hidden layer perceptron with ELU activation function and ten neurons in each hidden layer. To magnify the nonlinearity, we scaled the weights of the linear projections by a factor of $2.5$ on the basis of the default initialization in \verb|PyTorch|, which is a uniform distribution between $[-\frac{1}{\sqrt{n_{\text{in}}}}, \frac{1}{\sqrt{n_{\text{in}}}}]$. Bias terms were set to zero for simplicity. {Besides}, we added another exponential damping layer $e(x) = \exp(\frac{-||x||_{2}}{5})x$ on top of the perceptron, so that the deformation shrank to zero at the infinity. 

After obtaining $g$, we trained another DFORM network $\varphi$ between $f$ and $g$. $\varphi$ was parameterized as described in section~\ref{sec:NeuralODE}, except that it did not include the linear layer. The model was trained for 10000 batches {of 32 samples in a similar manner to the previous example}. For comparison, we also trained another DFORM network $\varphi_{l}$ with only the linear layer, using the same learning rate, batch size and number of training batches. After training, we evaluated the orbital similarity between $f$ and $g$ using the learned transformations {as in Eq.(\ref{eq:similarity})}. We also extracted the final orbital similarity loss ($l_{1} + l_{2}$, see Appendix~\ref{app:sampling}) after smoothing with an exponential kernel. We performed 30 experiments with different random systems. Each DFORM model was trained with five different initializations and the one with highest similarity was retained.

Results are shown in Figure~\ref{fig:param-rec}B-C. Consistent with previous examples, linear DFORM already provided a reasonable match, with similarity above 0.9, but the nonlinear method was indeed significantly better, both in the sense of lower loss (one-sided t-test, {$t(29) = -7.968$}, $p < 0.001$) and higher similarity (one-sided t-test, {$t(29) = 10.258$}, $p < 0.001$). One example was shown in Figure~\ref{fig:param-rec}C. While the difference in orbital similarity was small, the trajectory matching quality was higher with the nonlinear method. {Further, the pushforward distribution with the nonlinear DFORM also captured the `stretching' in the ground truth pushforward distribution better.}

\subsubsection{{Comparing bifurcating nonlinear systems through orbital similarity}} \label{sec:BLA}

{Having established that DFORM can align systems related by nonlinear transformations, here we proceed to explore the orbital similarity between topologically distinct systems. It is well-known that topologically distinct systems could be functionally similar, particularly when they are located on different sides of a subtle bifurcation. For example, \parencite{sagodiBackContinuousAttractor2024} showed that a bounded line attractor network sitting on a bifurcation point behaved very similarly with perturbed networks that possess monostable or multistable dynamics. We hypothesized that the orbital similarity between these systems could reflect their functional similarity. To test this hypothesis, we selected the line attractor system (no perturbation), monostable system (with perturbation weights $(0.1, 0)$; see Appendix~\ref{app:BLA} for details) and multistable system (with weights $(0, 0.1)$) from \parencite{sagodiBackContinuousAttractor2024}, as well as two systems that were further into the monostable (with weights $(0.3, 0)$) and multistable (with weights $(0, 0.3)$) regime. The vector fields and trajectories of these systems were visualized in Supplementary Figure~\ref{fig:BLA-sys}.}

\begin{figure}[htbp]

\begin{center}
\includegraphics[width=0.7\textwidth]{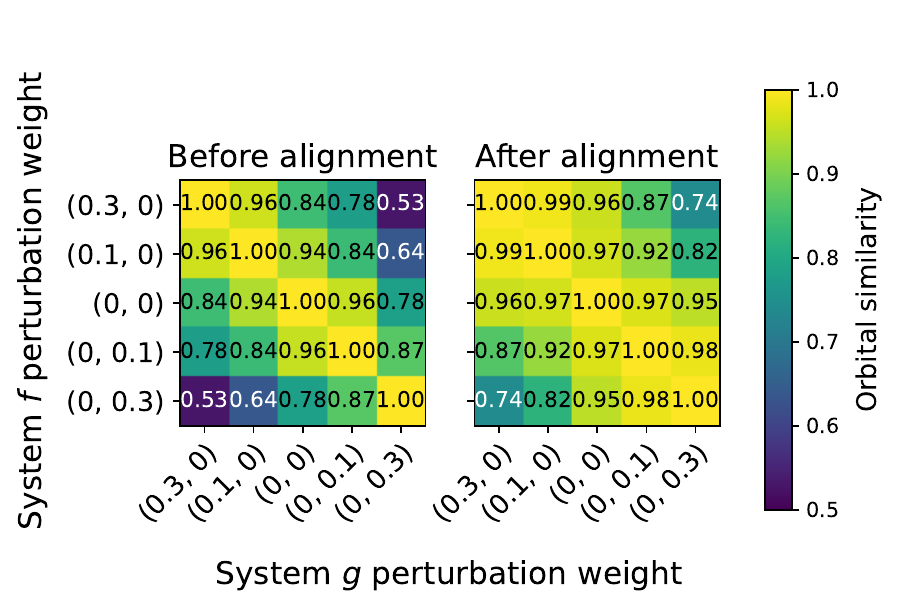}
\end{center}
\caption{{\bf {Similarity between bifurcating systems before and after nonlinear alignment.}} {Left: Similarity between the five selected systems without coordinate transformation, namely replacing $\varphi$ with the identity mapping in Eq.(\ref{eq:similarity}). Right: orbital similarity after nonlinear DFORM alignment. Note that the matrices were symmetric.}}
\label{fig:BLA-sim}
\end{figure}

{We trained a nonlinear DFORM between each pair of these systems. Similar to the Van der Pol oscillator example, we used 2000 batches of linear pretraining and 3000 batches of full-scale training, with a batch size of 32. Samples were drawn from a uniform distribution between -0.5 and 2 in both dimensions for all systems, which covers the limit sets of interest. The model with the best orbital similarity was selected out of five initializations. Results were shown in Figure~\ref{fig:BLA-sim}. Compared to the similarity matrix before alignment (i.e., replacing $\varphi$ in Eq.(\ref{eq:similarity}) with the identity mapping), the orbital similarity after nonlinear alignment showed clearer block structure. The similarity within groups of topologically congruent systems (system one and two; four and five) was almost perfect, and the similarity was much lower between groups. However, interestingly, the structurally unstable line attractor system (system three) was quite similar to both groups, consistent with their functional similarity. In summary, orbital similarity integrates both geometrical similarity and topological congruency, and provides a reasonable measure of functional similarity between dynamical models.}

\section{Locating dynamical motifs on low-dimensional manifolds}

In previous sections, we introduced the problem of diffeomorphic vector field alignment and demonstrated DFORM's capability of solving such problems. One major limitation of such a formulation is that the two systems under comparison must have the same dimensionality, because a diffeomorphism could not exist between manifolds of different dimensions. However, this condition might be too restrictive in practice.
In this section, we extend the DFORM framework to enable cross-dimensionality vector field alignment. We conduct experiments that demonstrate how this technique can help identify limit sets from a high-dimensional model. We then apply this technique to models fit on experimental neural data to reveal the low-dimensional features in the dynamics.

\subsection{Analyzing dynamical systems through DFORM template matching} \label{sec:cross_dim}

Here, we present two problems frequently encountered in the analysis of dynamical models and explained how to apply DFORM to tackle these problems.

\subsubsection{Triangular decomposition of nonlinear systems}

First, we consider the classical problem of identifying a \emph{triangular decomposition} of a nonlinear system \parencite{isidoriNonlinearControlSystems1995}. Given a system $f: \mathbb{R}^{n} \to \mathbb{R}^{n}$, we aim to find a diffeomorphism $\varphi \in \text{Diff}(\mathbb{R}^{n})$ such that the new coordinate $\xi = \varphi(x)$ can be split into two parts $\xi = (y, z)$ where $y \in \mathbb{R}^{m}$, $z\in\mathbb{R}^{n - m}$, and the pushforward vector field $\dot{\xi} = \varphi_{*}f(\xi)$ satisfies
\begin{equation} \label{eq:triangle}
\begin{cases}
    \dot{y} = g(y) \\
    \dot{z} = h(y, z)\,.
\end{cases}
\end{equation}

Triangular decomposition is useful because points with the same $y$-coordinate remain to have the same $y$-coordinate under the flow of $f$.
{While it is rarely possible to obtain an exact triangular decomposition with $m < n$, in the neuroscience setting, an approximal triangular decomposition could still be very useful. In this case, $\dot{y} = g(y) + \varepsilon(y, z)$ with $\|\varepsilon(y, z)\|_2 \ll \|g(y)\|_2$. We can try to identify an approximal triangular decomposition}
where $\dot{z} = h(y, z)$ represents some transient dynamics (with $z$ quickly converging to its stationary value). If so, the asymptotic behavior of the system will be governed by the lower-dimensional system $\dot{y} = g(y)$ that is easier to analyze.

DFORM provides a way to identify such triangularization. Given $f$, we can generate multiple hypotheses for its asymptotic dynamics $g(y)$ based on prior knowledge, numerical simulation or visualization {(see Discussion)}. We refer to these $g(y)$ as template `dynamical motifs'. Note that these templates need not to be geometrically similar to the actual dynamics - they just need to be topologically consistent. To evaluate the matching between low-dimensional templates and the high-dimensional system, we use DFORM to learn the diffeomorphism $\varphi$ and compute the difference between the pushforward $\varphi_{*}f$ and target $g$ in only the first $m$ dimensions.

Technically, denote by $T_{m}^{n}(v)$ the truncation of a vector $v$ to the first $m$ dimensions, and denote by $P_{n}^{m}(v)$ the $n$-dimensional vector obtained by padding $v \in \mathbb{R}^{m}$ with zeros. The new orbital similarity loss is thus defined as
\begin{equation} \label{eq:cross_dim}
    J_{f, g, \varphi}(\xi) = \frac{1}{m}\left\| \frac{T_{m}^{n}(\varphi_{*}f(\xi))}{\|T_{m}^{n}(\varphi_{*}f(\xi))\|_{2}} - \frac{g(T_{m}^{n}(\xi))}{\|g(T_{m}^{n}(\xi))\|_{2}}  \right\|_{2}^{2}  
\end{equation}
for $\xi\in\mathbb{R}^{n}$. If this loss vanishes for all $\xi = \varphi(x)$,  the first $m$ components of the pushforward vector field $\varphi_{*}f$ will align with the template $g$ and thus transform $f$ into the form of Eq~(\ref{eq:triangle}).

In practice, one might seek a decomposition that is accurate across some regions-of-interest rather than the whole phase space. Therefore, we can define the following loss term:
\begin{equation}
    l_{3} = \mathbf{E}_{x}\left[J_{f, g, \varphi}\big(\varphi(x)\big)\right]
\end{equation}
for $x \sim p_{x}$, where $p_{x}$ is a distribution over the regions-of-interest for system $f$.

\subsubsection{Identification of invariant submanifolds}

We now consider another motivating example. It is believed that neural dynamics exist on a manifold with much lower dimension than the full state space \parencite{langdonUnifyingPerspectiveNeural2023}. Many methods have been developed to discover such manifolds, broadly referred to as manifold learning or nonlinear dimensionality reduction. When the object of interest is a dynamical system, we can develop a new approach to this problem through the lens of invariant submanifolds. Technically, we assume that there exists an $m$-dimensional submanifold $\mathcal{M}$ of the $n$-dimensional phase space (with $m \ll n$), such that $\mathcal{M}$ is invariant under the flow of $f$, i.e., trajectories that starts in $\mathcal{M}$ stays in $\mathcal{M}$. If further, that the dynamics away from $\mathcal{M}$ are fast/transient and the dynamics on $\mathcal{M}$ are slow/persistent, then the state of the system will remain close to this manifold asymptotically. However, it is not easy to identify an invariant submanifold from a very high-dimensional space, particularly when the manifold is curved (in the Euclidean coordinates).

DFORM provides a way to identify such invariant submanifold through template matching. The idea is to find $\mathcal{M}$ such that (1) $\mathcal{M}$ is invariant under the flow of $f$; and that (2) $f$ restricted to $\mathcal{M}$ (denoted as $f|_{\mathcal{M}}$) is equivalent to an $m$-dimensional template $g$. To this aim, we learn a diffeomorphism $\varphi\in\text{Diff}(\mathbb{R}^{n})$ and define $\mathcal{M}$ as the $m$-dimensional slice in the new coordinate system $\varphi(x) = \xi = (y, z)$ with all $z$-coordinates being zero, i.e., $\mathcal{M} := \{\varphi^{-1}\big((y, \mathbf{0}_{n - m})\big)\mid y \in \mathbb{R}^{m}\}$. $\mathcal{M}$ is `flat' in the new coordinates $\xi$ but could be curved in the Euclidean coordinates $x$.

To formulate the optimization problem, we can `pad' $g$ to an $n$-dimensional vector field $\tilde{g}$. Let $\tilde{g} = P_{n}^{m}\circ g\circ T_{m}^{n}$. One can think of $\tilde{g}$ as `stacking' $m$-dimensional hyperplanes (with last $n - m$ coordinates being zero) where each hyperplane contains a replicate of $g$. The goal is to have $\varphi_{*}f$ aligned with $\tilde{g}$ for all points over the hyperplane $\{(y, \mathbf{0}_{n - m})\mid y\in\mathbb{R}^{m}\}$, or equivalently, $f$ aligned with $(\varphi^{-1})_{*}\tilde{g}$ for points over $\mathcal{M}$. We can thus define another orbital similarity loss term using the inverse of $\varphi$: 
\begin{equation}\label{eq:cross_dim_inv}
    J_{g, f, \varphi^{-1}}(x) = \frac{1}{n}\left\| \frac{(\varphi^{-1})_{*}\tilde{g}(x)}{\left\|(\varphi^{-1})_{*}\tilde{g}(x)\right\|_2} - \frac{f(x)}{\|f(x)\|_{2}} \right\|_2^2\,,
\end{equation}
where
\begin{equation}
    (\varphi^{-1})_{*}\tilde{g}(x) = \left(\frac{\partial \varphi}{\partial x}\bigg\rvert_{x}\right)^{-1}\tilde{g}(\varphi(x)) = \left(\frac{\partial \varphi}{\partial x}\bigg\rvert_{x}\right)^{-1}P_{n}^{m}(g(T_{m}^{n}(\varphi(x))))\,.
\end{equation}

If the loss vanishes for all points on $\mathcal{M}$, then $\mathcal{M}$ will be invariant with $f|_{\mathcal{M}}$ equivalent to $g$.

In practice, we can evaluate the loss by sampling points on $\mathcal{M}$ through $x = \varphi^{-1}\big((y, \mathbf{0}_{n - m})\big)$ with $y\in\mathbb{R}^{m}$. Therefore, we can define the following loss term:
\begin{equation}
    l_{4} = \mathbf{E}_{y}\left[J_{g, f, \varphi^{-1}}\big(\varphi^{-1}(P_{m}^{n}(y))\big)\right]
\end{equation}
for $y \sim p_{y}$, where $p_{y}$ is a distribution over the phase space of system $g$.

\subsubsection{DFORM implementation}

One may note that the two new loss functions Eq.(\ref{eq:cross_dim})  and Eq.(\ref{eq:cross_dim_inv}) resemble the orbital similarity loss defined in Section~\ref{sec:DFORM}, albeit with some extra truncation and padding operations. In fact, Eq.(\ref{eq:cross_dim}) and Eq.(\ref{eq:cross_dim_inv}) degenerate into Eq.(\ref{eq:topoloss}) and Eq.(\ref{eq:topoloss_inv}) when $n = m$. Therefore, we can utilize the same DFORM architecture to model and optimize the diffeomorphism $\varphi$. The model configuration and training method are the same as described in Section~\ref{sec:NeuralODE} except that the loss are computed differently (see Appendix~\ref{app:crossdim} for more details).

In the sections below, we present two toy examples to demonstrate how DFORM template matching can facilitate the analysis of high-dimensional models, and proceed to applying it on models trained on experimental neural data.

\subsection{\texorpdfstring{Recovery of low-dimensional invariant dynamics embedded in \\ high-dimensional systems}{Recovery of low-dimensional invariant dynamics embedded in high-dimensional systems}} \label{sec:SNIC}

As a first example, we showed that DFORM can identify a low-dimensional invariant manifold in a high-dimensional system, and locate the fixed points on this manifold. We constructed such a system in the following way: First, we generated a two-dimensional template for the Saddle Node on Invariant Cycle (SNIC) bifurcation, with bifurcation parameter $\mu = 0.5$ (see Appendix~\ref{app:SNIC}). This system has an invariant cycle that coincides with the unit circle, and two pairs of symmetric fixed points on the invariant cycle. One stable fixed point was located at $(\frac{\sqrt{3}}{2}, \frac{1}{2})$, and one saddle fixed point was located at $(-\frac{\sqrt{3}}{2}, \frac{1}{2})$. We then appended this two-dimensional system with a 16-dimensional RNN, which had monostable dynamics (see Appendix~\ref{app:RNN}). The combined system thus had independent dynamics in the first two coordinates and the last 16 coordinates, with one invariant hyperplane $\{(x, y, 0, \dots, 0)\mid x, y\in\mathbb{R}\}$. After that, we applied an 18-dimensional random orthogonal transformation $O$ to the combined system to mix all dimensions. The invariant hyperplane of the system will thus be $\mathcal{M}:\{w_{1}O_{1} + w_{2}O_{2}\mid w_{1}, w_{2}\in\mathbb{R}\}$, where $O_{i}$ refers to the $i$-th column of $O$.

We then tried to identify the invariant manifold $\mathcal{M}$ and the associated fixed points by aligning the 18-dimensional system $f$ back to the SNIC template $g$. We trained a linear DFORM between $f$ and $g$ for 20,000 batches of 32 samples. DFORM was able to achieve above 0.9 similarity between the two-dimensional projection of transformed vector field $\varphi_{*}f$ and $g$, indicating that the dynamics of the transformed system $\varphi_{*}f$ in the first two coordinates were well aligned to the template $g$.

\begin{figure}[htbp]

\begin{center}
\includegraphics[width=0.8\textwidth]{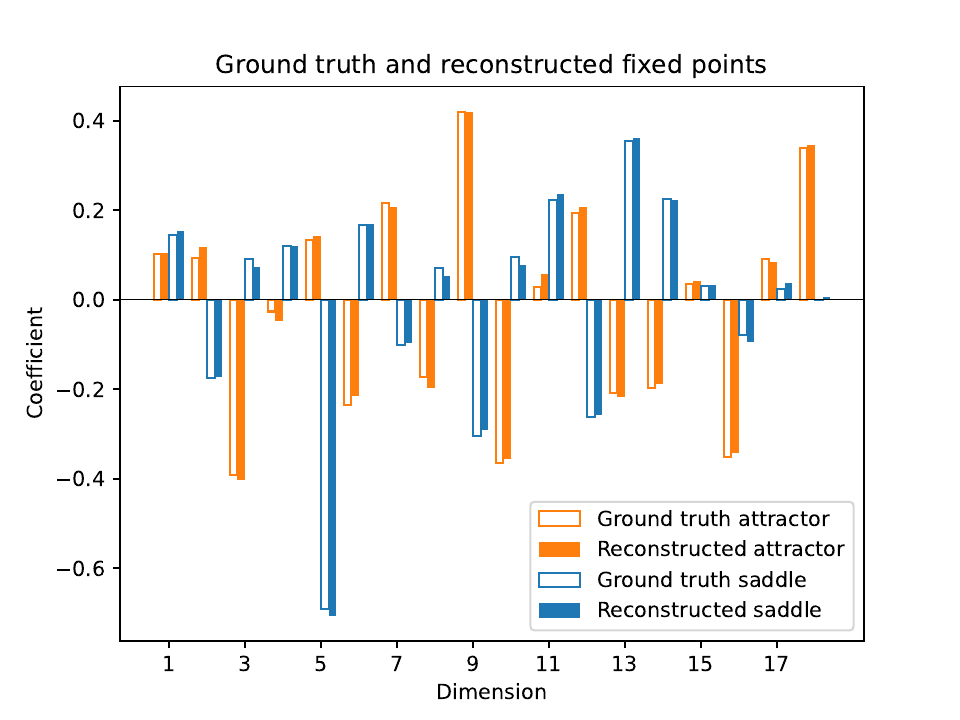}
\end{center}
\caption{{\bf Identification of saddle and attractive fixed points.} Bar plot showed the ground truth and reconstructed position of the attractive and saddle fixed points. Note that the system was symmetric so we did not show the other attractive and saddle fixed points that were the reflection of the ones shown. The position of each fixed point was given by an 18-dimensional vector, whose elements were visualized in the bar plot.}
\label{fig:SNIC}
\end{figure}

We then checked whether DFORM aligned the fixed points of the systems. Denote $x_{1} = (\frac{\sqrt{3}}{2}, \frac{1}{2}, 0, \dots, 0)$ and $x_{2} = (-\frac{\sqrt{3}}{2}, \frac{1}{2}, 0, \dots, 0)$, the ground truth positions of the stable and saddle fixed points of system $f$ were given by $Ox_{1}$ and $Ox_{2}$ respectively. Here, we utilized DFORM to find such fixed points by transforming the zero-padded fixed points in the template back into $f$'s phase space, namely $\varphi^{-1}(x_{1})$ and $\varphi^{-1}(x_{2})$ respectively. In Figure~\ref{fig:SNIC}, we visualized the elements of $Ox_{1}$, $Ox_{2}$, $\varphi^{-1}(x_{1})$, and $\varphi^{-1}(x_{2})$. The DFORM-reconstructed fixed points were very close to their correct position. Also note that because in this example the invariant manifold was a two-dimensional hyperplane and the transformations were linear, the fact that both fixed points were aligned well implies that the whole invariant hyperplane were aligned well, as any point on this hyperplane can be represented as a unique weighted sum of $Ox_{1}$ and $Ox_{2}$. Overall, these findings suggested that DFORM can be used as a tool to locate low-dimensional dynamical features from a high-dimensional system.

\subsection{Identification of a saddle limit cycle from high-dimensional system} \label{sec:saddle-cycle}

One of the most important advantages of DFORM template matching method is that it does not rely on numerical integration of the flow. Therefore, it is particularly suitable for identifying saddle limit sets, which cannot be found by simulating random trajectories either forward or backward in time.

\begin{figure}[htbp]

\begin{center}
\includegraphics[width=\textwidth]{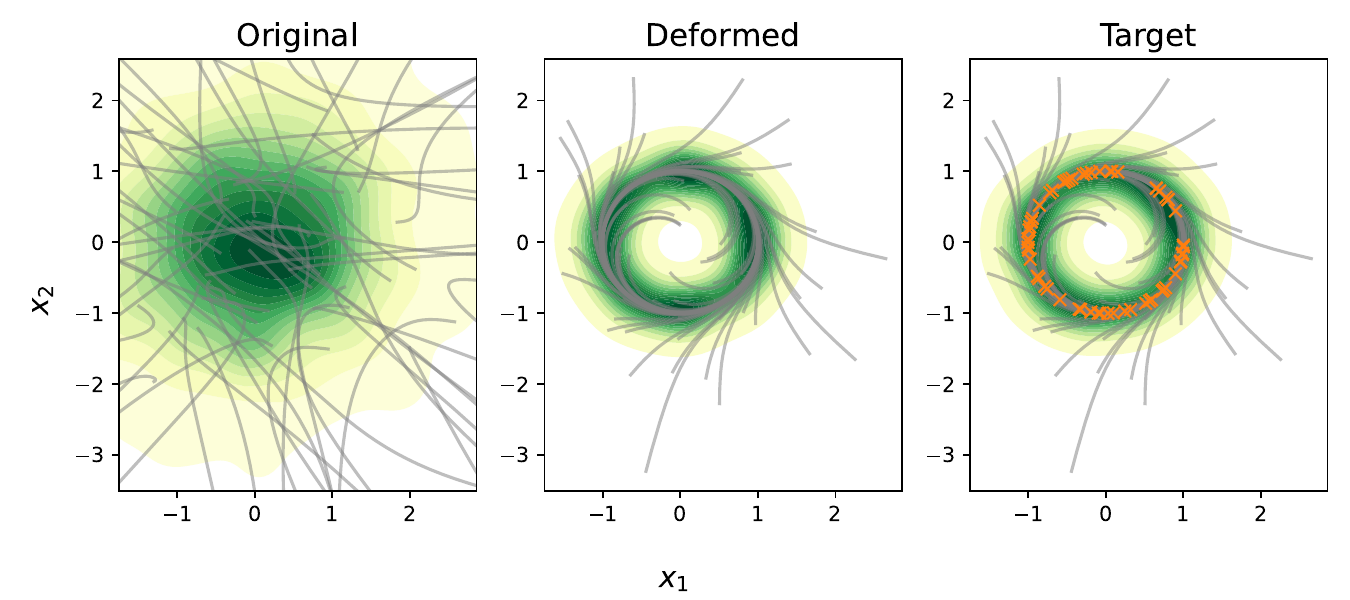}
\end{center}
\caption{{\bf Identification of a saddle limit cycle.} The original, transformed and target systems were visualized similar to Figure~\ref{fig:param-rec}, except that trajectories and distributions were projected directly to the first two coordinates of each system instead of the first two principal components. {Green contour plots indicate pushforward distributions. Gray lines represent simulated trajectories with orange crosses indicating endpoints of trajectories.}}
\label{fig:saddle-cycle}
\end{figure}

Here, we show that through matching to a two-dimensional limit cycle template, DFORM is able to identify a saddle limit cycle in a high-dimensional system. We constructed such a system in a way similar to the previous section: First, we designed a system composed of two independent parts. The dynamics in the first two dimensions of the system were given by the canonical form of Supercritical Hopf bifurcation with bifurcation parameter $\mu = 1$ (see Appendix~\ref{app:Hopf}), which contains a globally stable limit cycle (the unit circle) and an unstable fixed point (the origin). The dynamics in the last 16 dimensions were given by a random linear system with eight real eigenvalues and eight negative eigenvalues. The combined system will thus possess a single saddle fixed point at the origin, and a saddle limit cycle, with no attractors or repellers. Then, we applied an 18-dimensional orthogonal transformation $O$ to the combined system to intermix all dimensions, so that the saddle limit cycle will have nonzero loading in all coordinates. As the system did not contain any attractor or repeller, visualization of simulated trajectories did not reveal any obvious structure (Figure~\ref{fig:saddle-cycle}, left).

We then trained a DFORM model to align the first two dimensions of the system back to the Hopf bifurcation template. Training converged in around 1000 batches of 32 samples. DFORM achieved near perfect orbital similarity, and the saddle limit cycle was easily identifiable from the visualization of the projection of the trajectories onto the first two new coordinates (Figure~\ref{fig:saddle-cycle}, middle), which matched the template almost perfectly (Figure~\ref{fig:saddle-cycle}, right). Overall, this minimal example demonstrated the exciting potential of DFORM for identifying saddle limit sets in high-dimensional systems.

\subsection{Identification of dynamical features from resting state brain activity} \label{sec:MINDy_template}

In previous sections, we demonstrated DFORM's ability to compare dynamics and low-dimensional features across models using canonical examples and ground-truth simulations. Finally, we apply DFORM to a set of high-dimensional models fit on empirical neural data. We obtained 30 Mesoscale Individualized NeuroDynamics (MINDy) models from \parencite{singhEstimationValidationIndividualized2020}. These models were trained to approximate the functional magnetic resonance imaging (fMRI) recordings of individual human participants during resting state in the Human Connectome Project (HCP) dataset \parencite{vanessenWUMinnHumanConnectome2013}. Each model contained 100 interconnected units representing 100 brain parcels, with a sigmoidal activation function. Using numerical simulations, our previous study found that these models showed a wide range of dynamics, including multistability and stable limit cycles \parencite{chenDynamicalModelsReveal2025}. {Examples were visualized in Supplementary Figure~\ref{fig:MINDy-traj}.} However, a quantitative summary of the low-dimensional dynamical motifs in the models is lacking.

Here, we tried to provide a quantitative description by aligning the MINDy models to the Hopf bifurcation stable limit cycle template. The template was given by the normal form with $\mu = 1$ (same as in the previous section). We trained a linear DFORM model between each MINDy model $f$ and the template $g$. To facilitate sampling around the limit sets of the models, we drew samples from the asymptotic distribution of hidden states of both the MINDy model and the template, using a homogeneous Gaussian noise of standard deviation {0.05}. Models were trained for 3000 batches of 128 samples, with two different random initializations (the better one was selected). We quantified the cosine similarity between the projection of the transformed vector field to the first two coordinates and the target vector field{, namely, $\textbf{E}_{x}\left[ \cos\angle(T_{m}^{n}(\varphi_{*}f(x)), g(T_{m}^{n}(\varphi(x))) \right]$.}

\begin{figure}[htbp]

\begin{center}
\includegraphics[width=\textwidth]{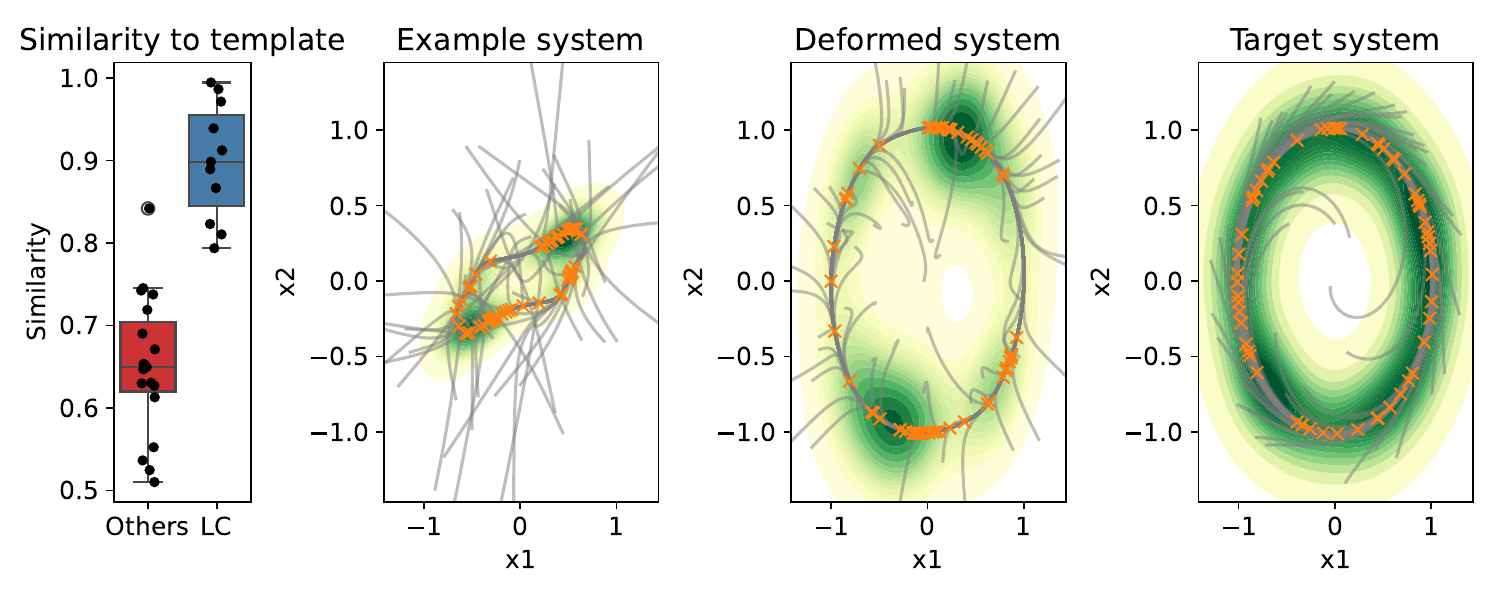}
\end{center}
\caption{{\bf Identification of limit cycles from empirical models of the resting brain.} Left panel: box and strip plots for the orbital similarity to the template after training for each MINDy model. Models were categorized according to the existence of stable limit cycles (`LC', blue box on the right) or not (`Others', red box on the left). Dots represent individual models. Other panels: the dynamics of an example model and its transformation to match the template. Same convention as Figure~\ref{fig:saddle-cycle}. {Green contour plots indicate pushforward distributions. Gray lines represent simulated trajectories with orange crosses indicating endpoints of trajectories.}}
\label{fig:MINDy}
\end{figure}

As expected, DFORM similarity measure agreed well with numerical simulations, showing a high value for systems that possess a stable limit cycle and a low value for the others, with a sharp boundary at around 0.8 (Figure~\ref{fig:MINDy} left). Interestingly, some models without a stable limit cycle also showed high similarity to the template, indicating the existence of {geometrically similar structure}. Visualization confirmed that the limit cycles were matched to the template pretty well (Figure~\ref{fig:MINDy} right). Overall, results suggested that DFORM can be used to identify low-dimensional dynamical features from high-dimensional models, and provide a more quantitative description compared to numerical simulations.

\section*{Conclusion and Discussion}

In this paper, we provide three major contributions. First, we developed DFORM, a computational framework for nonlinear vector field alignment through directly optimizing the diffeomorphism, without the need of linearization or kernel trick. Second, we validated the effectiveness of DFORM for nonlinear alignment and comparison of high-dimensional dynamical systems. Third, we demonstrated that DFORM can locate low-dimensional motifs within high-dimensional dynamics, providing a comprehensive approach for model analysis, manifold learning, and dimensionality reduction.

Recently, several methods have been developed for the alignment and comparison between dynamical models. However, these methods are generally linear in nature, with the extension to nonlinear mapping relying on basis function expansion or other linearization techniques. In this paper, we demonstrated that linearization-based alignment is insufficient for nonlinear systems, particularly for those with nontrivial limit sets (e.g., nonzero fixed points). On the contrary, DFORM adopts a fundamentally different perspective of directly optimizing over the group of nonlinear diffeomorphisms, which can be constrained to linear transformations too if that is desired. This technique thus allows us to directly compare the outcome of linear and nonlinear alignment. We found, reassuringly, that linear mapping was usually good enough for first-level approximation, providing reasonable alignment even when the ground truth transformation is nonlinear. However, linear mapping also obviously fell short in many cases illustrated in the paper, such as aligning two topologically equivalent linear systems or Van der Pol oscillators. We argue that such differences are crucial. Linear equivalency, while interesting, remains a limiting case of topological equivalency. DFORM's ability to identify a much wider set of nonlinear transformations will unlock a broad range of new possibilities, connecting common mechanisms across models despite highly heterogeneous coordinate systems.

One immediate application of DFORM is the exact linearization of nonlinear systems. The {extended Hartman-Grobman theorem \parencite{lanLinearizationLargeNonlinear2013}} states that a nonlinear system is topologically equivalent to its linearized system within the whole basin of attraction of a stable equilibrium. However, it is very hard to identify the homeomorphism between the two systems, preventing researchers from applying this powerful theorem to the numerical analysis of dynamical models. DFORM now provides a way to approximate such transformations and thus allow for linearization within the entire basin of attraction. If so, we can apply the rigorous analytical tools from linear systems theory to nonlinear dynamical models, representing a significant advance in the analysis of such models.

{One key component of the DFORM framework that we have not discussed extensively is the selection of sample distributions $x\sim p_{x}$ and $y\sim p_{y}$. A straightforward choice is the standard normal distribution or a uniform distribution over an area of interest, as in many of our experiments. However, for models that were fit on actual data, we recommend using a sample distribution that closely resembles the distribution of hidden states of the models in its typical operating condition. For example, $p_{x}$ can be the asymptotic distribution of the state $x$ when simulating the model $f$ under a certain amount of noise, as in our MINDy experiments. We also illustrated the effect of using different sample distributions in Appendix~\ref{app:distri}.}

{The current DFORM framework was designed to compare models by aligning trajectories, regardless of the speed discrepancy along these trajectories. However, DFORM can also be easily adapted to perform speed-sensitive alignment, by simply replacing the normalized Euclidean distance metric in the orbital similarity loss Eq.(\ref{eq:topoloss}) to an unnormalized one. In this case, the loss will be minimized only if $f$ is diffeomorphic to $g$. We envision that this speed-sensitive mode (which has been implemented in the associated code) will be particularly useful for comparing models with features like continuous attractors and slow manifolds \parencite{sagodiBackContinuousAttractor2024}, where the magnitude of the vector field is functionally important.}

Apart from alignment between models of the same dimensionality, DFORM also provides a new way to analyze high-dimensional dynamical systems by locating low-dimensional dynamical motifs.
Depending on the configuration, DFORM can be used to {(1) approximate a triangular decomposition of a nonlinear system, and (2) search for invariant low-dimensional dynamics embedded in a high-dimensional system}. One immediate application of DFORM is to locate the saddle limit sets of learned dynamical models, as we demonstrated in Section~\ref{sec:saddle-cycle}. Such analysis will generate new insights into the mechanistic underpinning of the models beyond their asymptotic behavior (i.e., the attractors). For example, with DFORM, it might be possible to probe whether competition mediated by saddle points \parencite{wongRecurrentNetworkMechanism2006a} is implemented in high-dimensional models of decision-making process.

{One challenge in cross-dimensionality alignment using DFORM is the selection of the dynamical `template' $g(y)$. In some cases, $g(y)$ might simply be another learned model. For example, one might want to study whether a higher-order model of the experimental data captures essentially the same dynamics as a lower-order model. DFORM can be used to examine whether the higher-order model $f(x)$ admits a triangular decomposition that reduces it to the lower-order model $g(y)$, or whether $f(x)$ converges to an invariant manifold where its dynamics are equivalent to $g(y)$. In other cases, $g(y)$ can be specified from the computational hypothesis, such as a canonical decision-making module with saddle points \parencite{wongRecurrentNetworkMechanism2006a}. In the absence of prior knowledge, one can also design a repertoire of templates $g(y)$ based on visualizations of simulated trajectories. Such visualization can be obtained from various dimensionality reduction methods such as PCA or diffusion Laplacian eigenmaps \parencite{coifmanDiffusionMaps2006, singerVectorDiffusionMaps2012}.}

Methodologically, there are several directions that DFORM can be extended. The first direction is to learn the transformation directly from timeseries data instead of the generative models. In principle, DFORM optimization only requires samples of system states and associated velocities, which can be approximated using temporal difference of timeseries data. This idea has been explored in \parencite{friedmanCharacterizingNonlinearDynamics2025} and showed promising results. The second direction is to extend DFORM to controlled systems. It is worth noting that DFORM is already compatible with controlled systems under constant/piecewise-constant inputs (as they can be represented by autonomous/piecewise-autonomous vector fields), which is the typical setting in theoretical neuroscience. However, it is more challenging to handle time-varying inputs. We are currently actively investigating into this problem.
{Finally, it could be interesting to survey popular alignment and comparison methods and apply them to a same set of representative computational and theoretical models in neuroscience, in order to understand whether they produce qualitatively similar or very different results.}

In conclusion, DFORM represents a significant advance in the analysis of high-dimensional nonlinear models, and opens up exciting new venues for future research. We also hope that the emergence of techniques like DFORM will make nonlinear dynamical modeling a more accessible and more powerful tool for scientific research. 

\section*{Code availability}

DFORM was implemented in Python and the scripts are available at \href{https://github.com/rq-Chen/DFORM_stable}{\texorpdfstring{https://github.com/rq-Chen/\\DFORM\_stable}{https://github.com/rq-Chen/DFORM\_stable}}.

\section*{Acknowledgments}

This research is funded by the National Institute of Mental Health 5R21MH132240-02 and the National Institute of Neurological Disorders and Stroke 5R01NS130693-03.
The views expressed are those of the authors and do not necessarily represent the official views of the National Institutes of Health.

\clearpage

\begin{appendices}

\section{Details of model optimization}

\subsection{Evaluation of the loss function} \label{app:sampling}

To ensure that the learned transformation $\varphi$ aligns the regions-of-interest in both $f$ and $g$ well, we combined four different variations of Eq.(\ref{eq:topoloss}) into our master loss function:
\begin{subequations} \label{eq:loss_terms}
\begin{align}
        l_{1} & = \mathbf{E}_{y}\left[J_{f, g, \varphi}(y)\right]\,, \\
        l_{2} & = \mathbf{E}_{x}\left[J_{g, f, \varphi^{-1}}(x)\right]\,, \\
        l_{3} & = \mathbf{E}_{x}\left[J_{f, g, \varphi}(\varphi(x))\right]\,, \label{eq:loss_term_3} \\ 
        l_{4} & = \mathbf{E}_{y}\left[J_{g, f, \varphi^{-1}}(\varphi^{-1}(y))\right]\,, \label{eq:loss_term_4}
\end{align}
\end{subequations}
where $x \sim p_{x}, y \sim p_{y}$. The first and third loss terms quantified the mismatch between $\varphi_{*}f$ and $g$, while the second and fourth ones quantified the mismatch between $f$ and $(\varphi^{-1})_{*}g$. The first two terms took the expected value by sampling from the codomain of the transformation ($y \sim p_{y}$ and $x \sim p_{x}$ for $\varphi$ and $\varphi^{-1}$ respectively), and the last two terms took the expected value by sampling from the domain of the transformation instead.

We also regulated the magnitude of the nonlinear transformation, given by
\begin{equation}
\|v\|_{H}^{2} := \textbf{E}_{x}\left[ \|v(x)\|_{2}^{2}\right]\,,
\end{equation}
where $x$ follows some probability distribution over $\mathbb{R}^{n}$, and $v$ represents the Neural ODE network. Here we chose the standard normal distribution with identity covariance. We evaluated this expected value by taking the average of $||v(x)||_{2}^{2}$ over 128 random samples. The regularization term for the orthogonality of the linear transformation is given by the {squared} Frobenius norm $\|HH^{T} - I_{n}\|_{F}^{2}$, where $H$ is the weight matrix of the linear layer. {The regularizers were normalized by the dimension of the model as in the orbital similarity loss.}

The final loss function is thus
\begin{equation}
    l = \lambda_{1}l_{1} + \lambda_{2}l_{2} + \lambda_{3}l_{3} + \lambda_{4}l_{4} + \frac{\lambda_{5}}{n}\|v\|_{H}^{2} + \frac{\lambda_{6}}{n}\|HH^{T} - I_{n}\|_{F}^{2}\,,
\end{equation}
where $\lambda_{i}, i = 1, 2, \dots, 6$ are hyperparameters. Unless stated otherwise, we let $\lambda_{1} = \lambda_{2} = 1$, $\lambda_{3} = \lambda_{4} = 0$ (equal weights for the forward and inverse transformation, and only sampling from codomain) and $\lambda_{5} = \lambda_{6} = 10^{-3}$ in all experiments where $f, g$ have the same dimensionality. {Such hyperparameter choice is well-motivated. Weights for $l_{1}$ and $l_{2}$ should be equal and positive, because we aim to match the transformed source vector field $\varphi_{*}f$ to the target vector field $g$ in its `important' regions (characterized by $p_{y}$) and vice versa. On the contrary, terms $l_{3}$ and $l_{4}$ evaluated the loss not by the prespecified distributions $p_{y}$ and $p_{x}$, but by learned pushforward distributions $\varphi_{*}p_{x}$ and $(\varphi^{-1})_{*}p_{y}$. Consequently, to minimize $l_{3}$ or $l_{4}$, the optimization could `cheat' by learning an awkward pushforward distribution. For example, the pushforward distribution could be unreasonably `squeezed' to minimize discrepancy in hard-to-match dimensions; it can also be unreasonably extended such that the mismatch within bounded regions vanishes during averaging. Therefore, when the topological congruency between $f$ and $g$ is unknown, it is most reasonable to set $\lambda_{3}$ and $\lambda_{4}$ both to zero. Nonetheless, when $f$ and $g$ are indeed equivalent, we found that the solution was quite robust to any choice of $\lambda_{1}$ to $\lambda_{4}$.}

\subsection{Cross-dimensionality matching} \label{app:crossdim}

In Section~\ref{sec:cross_dim}, we introduced two new loss terms:
\begin{subequations} \label{eq:master_loss_crossdim}
\begin{align}
    l_{3} &= \mathbf{E}_{x}\left[J_{f, g, \varphi}\big(\varphi(x)\big)\right]\,, \\
    l_{4} &= \mathbf{E}_{y}\left[J_{g, f, \varphi^{-1}}\big(\varphi^{-1}(P_{m}^{n}(y))\big)\right]\,,
\end{align}
\end{subequations}
which resembles Eq.(\ref{eq:loss_term_3}) and Eq.(\ref{eq:loss_term_4}) in the previous section.
In fact, we can naturally generalize Eq~(\ref{eq:loss_terms}) to the case where $m \le n$ as:
\begin{subequations} \label{eq:loss_terms_crossdim}
\begin{align}
        l_{1} & = \mathbf{E}_{y}\left[J_{f, g, \varphi}\big(P_{n}^{m}(y)\big)\right] \\
        l_{2} & = \mathbf{E}_{x}\left[J_{g, f, \varphi^{-1}}(x)\right] \\
        l_{3} & = \mathbf{E}_{x}\left[J_{f, g, \varphi}\big(\varphi(x)\big)\right] \\
        l_{4} & = \mathbf{E}_{y}\left[J_{g, f, \varphi^{-1}}\left(\varphi^{-1}\big(P_{n}^{m}(y)\big)\right)\right]
\end{align}
\end{subequations}
for $x \in \mathbb{R}^{n}$, $x \sim p_{x}$ and $y \in \mathbb{R}^{m}$, $y \sim p_{y}$. Note that Eq~(\ref{eq:loss_terms_crossdim}) degenerate into Eq~(\ref{eq:loss_terms}) when $n = m$.

When $m\ne n$, the differences between the four loss terms become more essential. By manipulating the hyperparameters $\lambda_{i}$, we can enforce different kinds of alignment. As described in Section~\ref{sec:cross_dim}, a high $\lambda_{3}$ is good for triangular decomposition and a high $\lambda_{4}$ is good for identification of invariant manifold.
{However, as described in the previous section, relying only on $l_{3}$ and $l_{4}$ could lead to unreasonable solutions. Therefore, we still include a positive weight for terms $l_{1}$ and $l_{2}$, which quantifies the quality of match by predefined measures.}
For the ground-truth recovery examples in this paper (section~\ref{sec:SNIC} and~\ref{sec:saddle-cycle}), we aimed to identify a triangular decomposition that simultaneously locates the invariant manifold, so we used bigger weights for both $\lambda_{3}$ and $\lambda_{4}$ ($\lambda_{1} = \lambda_{2} = 1$, $\lambda_{3} = \lambda_{4} = 10$). For the empirical examples (section~\ref{sec:MINDy_template}), we focused on the invariant dynamical features and thus used a much higher weight for $\lambda_{4}$ ($\lambda_{1} = 1$, $\lambda_{2} = \lambda_{3} = 10$, $\lambda_{4} = 100$).

\section{Details of the dynamical models}

\subsection{Linear systems}  \label{app:linear-systems}

A linear system is defined as $\dot{x} = f(x) = Ax$, where $x\in\mathbb{R}^{n}$ and $A\in\mathbb{R}^{n \times n}$. In this paper, we performed experiments using two types of linear systems: (1) random linear systems, where the entries of $A$ were drawn from a normal distribution with zero mean and variance $\frac{1}{n}$; and (2) linear systems with desired \emph{signature}. We define the signature of a matrix $A$ (and similarly the linear system whose dynamic matrix is $A$) as a triple $(p, q, r)$ where $p, q, r$ are the numbers of $A$'s eigenvalue with positive, negative and zero real parts, respectively. We first selected the absolute value of the real and imaginary parts of all eigenvalues from the uniform distribution over $[0, 1]$ and assigned signs to them according to the signature, to obtain the eigenspectrum of the matrix. We then constructed a block diagonal matrix $A_{0}$ where each block is either a $1\times 1$ matrix of a real eigenvalue, or a $2 \times 2$ matrix $\bigl(\begin{smallmatrix}a & -b \\ b & a\end{smallmatrix}\bigl)$ corresponding to the complex conjugate eigenvalues $a \pm bi$. After that, we applied a similar transformation to $A_{0}$ by a random orthogonal matrix $H$, and obtained the desired matrix $A = HA_{0}H^{-1}$. The matrix $H$ was obtained by first constructing a random matrix $H_{0}$ as described in (1), then applied the QR decomposition $H_{0} = HR$.

For all experiments in the paper using linear systems, the sample distribution was set to the multivariate normal distribution with zero mean and identity covariance matrix. In Section~\ref{sec:paramrec}, a nonlinear system $g$ was generated by pushing forward a linear system $f$ with a random DFORM model $\varphi_{0}$. The sample distribution for system $g$ was also set to the pushforward of the sample distribution of $f$ (the standard multivariate normal distribution) by $\varphi_{0}$. Technically, random samples of $y$ were generated by randomly sampling $f$ from the normal distribution{, and} then set $y = \varphi_{0}(x)$.

\subsection{RNNs} \label{app:RNN}

We built RNNs with `low-rank plus random' connectivity profile as described in \parencite{schuesslerDynamicsRandomRecurrent2020}. The dynamics of a `vanilla' RNN is given by $\dot{x} = -x + W\tanh(x)$, where $x\in\mathbb{R}^{n}$ and $W \in \mathbb{R}^{n\times n}$. Following \parencite{schuesslerDynamicsRandomRecurrent2020}, we constructed $W$ as $W = J + \sum_{i = 1}^{K} m_{i}n_{i}^{\top}$, where $J\in\mathbb{R}^{n \times n}$ is a random matrix, $m_{i}, n_{i}\in\mathbb{R}^{n}$ are vectors and $K \in\mathbb{N}$ is the desired rank of the low-rank connectivity component. The entries of $J$ were drawn from a normal distribution with zero mean and variance $\frac{g^{2}}{n}$, where $g$ is the expected value of the radius of the bulk of the eigenspectrum of $W$. Entries of $m_{i}$ were drawn from standard normal distribution. $n_{i}$ was constructed from $m_{i}$ in such a manner that the expected value of $n_{i}^{\top}J^{j}m_{i}$ for $j = 0, 1, \dots$ equals a designated value $\theta_{j}$, which leads to outliers in the eigenspectrum of $J$. For our experiment, we set $K = 1$, $g = 0.9$, $\theta_{0} = 1.5$, and $\theta_{j} = 0, \forall j > 0$. If $n$ is large enough, this algorithm leads to an eigenspectrum of $J$ that has a bulk part centered at the origin with radius 0.9, and one real eigenvalue around 1.5. The RNN will have a pair of symmetric non-zero stable fixed point with the origin being unstable. In our experiments, we forward simulated the models to identify their attractors. As our network size was much smaller than the thermodynamic limit, our RNNs mostly showed bistable or oscillatory (with unstable origin) or monostable (with stable origin) behavior. We were thus able to demonstrate the capability of DFORM over more types of nonlinear dynamics that were commonly observed in practice \parencite{chenDynamicalModelsReveal2025}.

To construct `linearly transformed' RNN systems, we defined a new system based on the definition of pushforward vector field. Given an $n$-dimensional RNN $\dot{x} = f(x)$ and a matrix $H\in\mathbb{R}^{n\times n}$, we defined the new system as $\dot{y} = g(y) = Hf(H^{-1}y) = -y + HW\tanh(H^{-1}y)$. Here, $H$ is a random matrix (orthogonal or non-orthogonal) constructed in the same way as in the previous section. The new system $g$ might not be in the form of a `vanilla' \parencite[Amari-type;][]{amariDynamicsPatternFormation1977a} RNN. However, note that when $H = W^{-1}$, $\dot{y} = -y + \tanh(Wy)$ is the Wilson-Cowan-type equivalent of the original Amari-type RNN \parencite{wilsonExcitatoryInhibitoryInteractions1972}. The sample distribution of $f$ was set to the multivariate normal distribution with zero mean and identity covariance, and the sample distribution of $g$ was set to the pushforward of this distribution by the transformation $\mathcal{H} = Hx$, i.e., a multivariate normal distribution with zero mean and covariance matrix $HH^{T}$.

\subsection{Low-dimensional dynamical templates}

\paragraph{Van der Pol oscillator.}  \label{app:VDP}
Dynamics of the Van der Pol oscillator system with bifurcation parameter $\mu \in\mathbb{R}$ is given by:
\begin{equation}
\begin{cases}
    \dot{x} = y \\
    \dot{y} = \mu(1 - x^{2})y - x\,.
\end{cases}
\end{equation}

When $\mu$ is negative or zero, the system possess a globally stable fixed point at the origin; when $\mu$ is larger than zero, the system possess a globally stable limit cycle of irregular shape, and an unstable fixed point at the origin.

In our experiments, we utilized the model with $\mu > 0$ for nonlinear alignment of limit cycles. Samples were drawn from a uniform distribution over the rectangle area with $x$-coordinate between $[-3, 3]$ and $y$-coordinate between $[-1.5\mu - 3, 1.5\mu + 3]$. Such sample distribution covers the limit cycle, as shown by the dark contour plot in the left and right panels in Figure~\ref{fig:VDP}.

\paragraph{{Bounded line attractor.}} \label{app:BLA}
{We defined the bounded line attractor system as in \parencite{sagodiBackContinuousAttractor2024}. The system is given by $\dot{x} = -x + \text{ReLU}(Wx + b)$ where $W = [\begin{smallmatrix}0 & -1\\ -1 & 0\end{smallmatrix}]$ and $b = [\begin{smallmatrix}1\\1\end{smallmatrix}]$. This system has a continuous line attractor. Under the perturbation $W \to W + w_1V_1$ where $w_{1}$ is small and $V_{1} = [\begin{smallmatrix}-2 & 1\\ 1 & -2\end{smallmatrix}]$, the system would bifurcate towards dynamics with only one fixed point (and it is stable). Under the perturbation $W \to W + w_2V_2$ where $V_{2} = [\begin{smallmatrix}1 & -2\\ -2 & 1\end{smallmatrix}]$, the system would bifurcate towards dynamics with one saddle point and two stable fixed points. In our experiment, we selected the systems with $(w_{1}, w_{2})$ being $(0.3, 0), (0.1, 0), (0, 0), (0, 0.1), (0, 0.3)$. The vector fields and simulated trajectories of these systems were visualized in Figure~\ref{fig:BLA-sys}.}

\paragraph{Saddle Node on Invariant Cycle (SNIC) bifurcation template.}  \label{app:SNIC}
We constructed a system that manifest SNIC bifurcation while being symmetric under reflection, mainly because the models to be compared against these templates (introduced below) are also symmetric. The dynamics of the SNIC bifurcation template are given in polar coordinates by:
\begin{equation}
    \begin{cases}
        \dot{r} = r(1 - r^{2})\\
        \dot{\theta} = \mu - |\sin\theta|\,,
    \end{cases}
\end{equation}
where $\mu\in\mathbb{R}$ is the bifurcation parameter. It is easy to see that this system has an unstable fixed point at the origin and an invariant cycle that coincides with the unit circle. When $0 < \mu < 1$, there exist a pair of stable fixed points on the invariant cycle, located at Euclidean coordinates $(\sqrt{1 - \mu^{2}}, \mu)$ and $(-\sqrt{1 - \mu^{2}}, -\mu)$ respectively; and a pair of saddle fixed points on the invariant cycle too, located at $(-\sqrt{1 - \mu^{2}}, \mu)$ and $(\sqrt{1 - \mu^{2}}, -\mu)$ respectively. When $\mu \ge 1$, the saddles merged with the attractors and there remains only one globally attractive limit cycle (the unit circle) and an unstable fixed point at the origin.

To facilitate alignment of the invariant cycle, we drew random samples $(r, \theta)$ from this system by choosing $r$ from a uniform distribution between $0.8$ and $1.2$, and choosing $\theta$ from a uniform distribution between $0$ and $2\pi$. In Section~\ref{sec:SNIC}, we constructed a high-dimensional system $f$ by combining the dynamics of the SNIC template $\dot{y} = g(y)$ and a random RNN $\dot{z} = h(z)$, then applying an orthogonal transformation $\mathcal{H}$. The samples of system $g$ were drawn from the pushforward of the joint distribution $(y, z)$ by $\mathcal{H}$, with $y$ and $z$ were independently drawn from the sample distributions for the SNIC template (described above) and the RNN (standard normal distribution) respectively.

\paragraph{Supercritical Hopf bifurcation normal form.}  \label{app:Hopf}
The normal form of the supercritical Hopf bifurcation is given by:
\begin{equation}
    \begin{cases}
        \dot{x} = (\mu - x^{2} - y^{2})x - y\\
        \dot{y} = x + (\mu - x^{2} - y^{2})y\,,
    \end{cases}
\end{equation}
which has a globally stable fixed point at the origin when $\mu \le 0$; and an unstable fixed point at the origin and a globally stable limit cycle $r = \sqrt{\mu}$ when $\mu > 0$.

In Section~\ref{sec:saddle-cycle}, we constructed a high-dimensional system $f$ containing a saddle limit cycles by combining a Hopf bifurcation template $\dot{y} = g(y)$ and a random linear system $\dot{z} = h(z)$, followed by an orthogonal transformation $\mathcal{H}$. Similar to Section~\ref{sec:SNIC}, the samples of system $f$ were also drawn from the pushforward of the joint distribution $(y, z)$ by $\mathcal{H}$. $y$ were drawn using polar coordinate $(r, \theta)$ with $r$ uniformly distributed between $[0.8\mu, 1.2\mu]$ and $\theta$ uniformly distributed between $[0, 2\pi]$, as visualized in the right panel in Figure~\ref{fig:saddle-cycle}. $z$ were drawn from the standard normal distribution independently from $y$.
In Section~\ref{sec:MINDy_template}, the samples were drawn using the asymptotic distribution of the states $y$ under a gaussian white noise of standard deviation {0.05} instead, as visualized in the right panel of Figure~\ref{fig:MINDy}.

\subsection{MINDy models}  \label{app:MINDy}

A Mesoscale Individualized NeuroDynamics (MINDy) model \parencite{singhEstimationValidationIndividualized2020, singhPrecisionDatadrivenModeling2025} is a neural mass model for whole-brain neural activity dynamics recorded by functional magnetic resonance imaging (fMRI) or Magneto/Electroencephalography (M/EEG). In this paper, we focused on the fMRI variant of MINDy. The dynamics of the model is given by
\begin{align}
    \dot{x} &= W{\psi}_{{\alpha}}(x) - D\odot x \\
    \psi_{\alpha}(x) &= \sqrt{\alpha^{2} + (bx + 0.5)^{2}} - \sqrt{\alpha^{2} + (bx - 0.5)^{2}}\,,
\end{align}
where $\mathbf{x}\in\mathbb{R}^{n}$ represents the hidden state of $n$ cortical regions. $W\in \mathbb{R}^{n\times n}$ is the effective connectivity matrix. ${\psi}_{{\alpha}}$ is an element-wise nonlinear activation function for each brain region, parameterized by curvature $\alpha > 0$ and slope $b > 0$ that differ across each dimension/region. $D \in \mathbb{R}^{n}$ is the decay rate for each region and $\odot$ indicates element-wise multiplication. The parameters $W, \alpha, D$ were fit directly from neural activation timeseries $x_{t}$ derived from fMRI recordings, such that the predicted difference $W\psi_{\alpha}(x_{t}) - D\odot x_{t}$ approximated the observed difference $x_{t+1} - x_{t}$. Parameter $b$ was fixed at $\frac{20}{3}$. In this paper, we adopted the MINDy models fit on resting-state fMRI recordings in the Human Connectome Project (HCP), as described in \parencite{singhEstimationValidationIndividualized2020}. Each model was fit on one individual's data and contained 100 brain regions (i.e., $n = 100$). These models were shown to display a variety of nonlinear dynamical landscapes, including monostability, mulitistability, oscillations, and the combinations of above \parencite{chenDynamicalModelsReveal2025}.
In our experiments, the sample distribution for MINDy models were chosen to be the asymptotic distribution of states under homogeneous Gaussian white noise with standard deviation {0.05}. Note that {MINDy was trained on standardized data with unit variance}.

\section{Non-identifiability of solutions} \label{app:identifiability}

Here, we aim to prove the non-identifiability of the diffeomorphic vector field alignment problem and demonstrate its implication for the alignment of nonlinear systems. The theory is independent of the optimization framework and also largely independent of the exact form of the loss function, thus equally applicable to DFORM and other methods like DSA \parencite{ostrowGeometryComparingTemporal2023}. We define non-identifiability as the property that given a vector field $f$, the map between transformations $\varphi$ and transformed systems $\varphi_{*}f$ is not one-to-one but many-to-one, thus preventing the recovery of `ground truth transformation'.

Assuming that we were given two systems $\dot{x} = f(x)$ and $\dot{y} = g(y)$ with $x, y \in \mathbb{R}^{n}$. We want to analyze the existence and uniqueness of the solution to the following optimization problem:
\begin{equation}
    \min_{\varphi\in G} \mathcal{L}(\varphi_{*}f, g)
\end{equation}
where $G\subseteq\text{Diff}(\mathbb{R}^{n})$ is a subgroup of the group of diffeomorphisms over $\mathbb{R}^{n}$ and $\mathcal{L}$ is a loss function characterizing the mismatch. Obviously, we have $\mathcal{L}(\varphi_{*}f, g) = \mathcal{L}(\psi_{*}f, g)$ if $\varphi_{*}f\equiv \psi_{*}f$. In the following sections, we will focus on this special case and discuss the conditions for it to hold. It turns out that {for general $G$} there always exists $\psi\ne\varphi$ with $\psi_{*}f = \varphi_{*}f$, proving the non-identifiability of the problem. We will also explain how this property matters when trying to align nonlinear systems with linearization-based method.

\subsection{Linear alignment between linear systems}

Given two linear systems $\dot{x} = f(x) = Ax$ and $\dot{y} = g(y) = By$. We consider a linear coordinate transformation $\mathcal{H}\in \text{GL}(n, \mathbb{R})$, namely $\mathcal{H}(x) = Hx$ and $\det H \ne 0$. The coordinate-transformed system is given by $\dot{y} = \mathcal{H}_{*}f(y) = HAH^{-1}y$. We can define a loss function as
\begin{equation}
    \mathcal{L}(\mathcal{H}_{*}f, g) := \|D\varphi_{*}f - Dg\|_{F} = \|HAH^{-1} - B\|_{F}
\end{equation}
which is the loss function for DSA \parencite{ostrowGeometryComparingTemporal2023}. Here $D$ represents the derivative/Jacobian operator. Note that we always use $D\varphi_{*}f$ to refer to $D(\varphi_{*}f)$ rather than $(D\varphi)_{*}f$.

Since $H_{1}, H_{2}$ are both invertible, we have
\begin{flalign}
&(\mathcal{H}_{1})_{*}f = (\mathcal{H}_{2})_{*}f \\
\iff & H_{1}AH_{1}^{-1}y = H_{2}AH_{2}^{-1}y, \forall y \in \mathbb{R}^{n} \\
\iff & H_{1}AH_{1}^{-1} = H_{2}AH_{2}^{-1} \\
\iff & (H_{1}^{-1}H_{2})A = A(H_{1}^{-1}H_{2}) \,,
\end{flalign}
which provides the following theorem:

\begin{Theorem} \label{th:linear}
Given a linear system $\dot{x} = f(x) = Ax$ with $A\in M_{n}(\mathbb{R})$ and an invertible linear transformation $\mathcal{H}_{1}: \mathcal{H}_{1}(x) = H_{1}x$, the linear transformation $\mathcal{H}_{2}: \mathcal{H}_{2}(x) = H_{2}x = H_{1}Jx$ is invertible and the pushforward systems $(\mathcal{H}_{1})_{*}f = (\mathcal{H}_{2})_{*}f$ if and only if $J = H_{1}^{-1}H_{2}$ is invertible and commutes with $A$.
\end{Theorem}

This theorem implies that the level set of the loss function containing $\mathcal{H}_{1}$ will also contain all $\mathcal{H}_{2}$ such that $H_{2} = H_{1}J$ with $J\in C(A)$ a commutant of $A$ and $J$ invertible. It is well-known that $C(A)$ is a linear subspace of $M_{n}(\mathbb{R})$ with at least $n$ dimension. Therefore, the level set will also contain a manifold of at least $n$ dimension, rather than only isolated points. To understand the structure of the level set better, consider two particularly interesting cases:

First, the case where $J = cI_{n}$ with $c\in\mathbb{R}, c\ne 0$, namely $H_{2} = cH_{1}$. This implies that the loss function is invariant under scaling. Therefore, instead of considering all invertible linear transformations, without loss of generality, we can restrict the optimization to matrices with unit Frobenius norm, which is a compact subset of $M_{n}(\mathbb{R})$. Assuming that the loss function $\mathcal{L}$ is continuous, the minimum of $\mathcal{L}$ must be attained over this compact set, proving the existence of solution. However, the optimal solution is obviously non-unique even when restricting to unit Frobenius norm.

Second, the case where $J = P(A)$ is a function of $A$, and particularly when $J = e^{TA}$ for $T \in \mathbb{R}$. In the latter case, the transformation $J$ maps every point $x$ to its position after time $T$ along the flow of the original system. Therefore, $J$ represents a temporal translation and thus will not alter the (time-invariant) vector field. Importantly, this case can be naturally generalized to nonlinear systems, as shown in later sections.

\subsection{Implication for linearization-based alignment of nonlinear systems} \label{app:linearization}

While non-identifiability of linear transformations might not be very important for alignment between linear systems, it does create difficulties for nonlinear systems. We consider the two following systems that were related by a linear transformation $y = \mathcal{H}(x) = Hx$:
\begin{align}
    f(x) &= Ax + \varepsilon(x) \\
    g(y) &= HAH^{-1}y +H\varepsilon(H^{-1}y)\,.
\end{align}

We assumed that the two systems are obtained through some linearization procedure, and $\varepsilon$ represents the residual error, whose value and Jacobian vanishes at the origin. For Jacobian-based linearization, $x$ and $y$ could be the original coordinate and $A$ being the Jacobian at the origin; for Koopman operator linearization, $x$ and $y$ could be the truncation of the infinite-dimensional Koopman eigenspectrum. After linearization, one would train a linear coordinate transformation between the two linearized systems:
\begin{align}
    \hat{f}(x) &= Ax \\
    \hat{g}(y) &= HAH^{-1}y\,.
\end{align}

As explained above, if the optimization is successful, one would identify a solution $\mathcal{H}_{2}(x) = H_{2}x$ where $H_{2} = HJ$ with $J$ being an invertible commutant of $A$. Because the optimal solution set contains an at least $n$-dimensional subspace, the probability of $H_{2}$ being close to the ground truth $H$ is basically zero. However, unlike the linear case where both $H$ and $H_{2}$ produce the same pushforward vector field, in the nonlinear case they are different. To see this, let's assume that there exists a nonzero equilibrium $x^{0}$ for system $f$, namely $Ax^{0} + \varepsilon(x^{0}) = \mathbf{0}$. If the alignment is good, we would expect $H_{2}x^{0}$ to be an equilibrium of system $g$, or at least close to being so. To check this hypothesis, we compute the derivative of system $g$ at $H_{2}x^{0}$ using Taylor expansion:
\begin{align}
    g(H_{2}x^{0}) & = HAH^{-1}HJx^{0} + H\varepsilon(H^{-1}HJx^{0}) \\
    & = HAJx^{0} + H\varepsilon(Jx^{0}) \\
    & = HAJx^{0} + H\left[ \varepsilon(x^{0}) + D\varepsilon|_{x^{0}}(J - I_{n})x^{0} + o\left(\|(J - I_{n})x^{0}\| \right) \right]\,.
\end{align}
Denote the Jacobian $D\varepsilon|_{x^{0}} = K$ and remove the higher-order term, we have:
\begin{align}
    g(H_{2}x^{0}) & \approx HAJx^{0} + H\left[ -Ax^{0} + K(J - I_{n})x^{0}  \right]\\
    & = HA(J - I_{n})x^{0} + HK(J - I_{n})x^{0} \\
    & = H(A + K)(J - I_{n})x^{0}\,.
\end{align}
which is not a small term unless $J \approx I_{n}$.

Therefore, linearization-based alignment will most likely mapped a nontrivial equilibrium of system $f$ to a non-equilibrium state of $g$, with $\|g(y)\|$ well above zero. As equilibria are usually the most important features of a dynamical model, this conclusion shows that linearization-based alignment is fundamentally limited for nonlinear systems.

\subsection{Orthogonal alignment}

While we derive our theory based on the general linear group, most existing works focused on the orthogonal or special orthogonal group \parencite{lipshutzDisentanglingRecurrentNeural2024, ostrowGeometryComparingTemporal2023}. The usage of the orthogonal group is well-motivated. While highly restrictive, the orthogonal group has an excellent property that it preserves distance and angle, which can induce a pseudo-metric over the space of vector fields \parencite{williamsGeneralizedShapeMetrics2021}. Also, when only considering orthogonal transformations, the multiplicity issue can also be significantly mitigated. That is because the commutant $J$ must also be an orthogonal matrix in order to make $H_{2} = HJ$ orthogonal, given a ground-truth orthogonal transformation $H$. In that case, the fact that $JA = AJ$ for some orthogonal $J$ implies that the system already has some rotational and reflective symmetry. While this could happen in some neuroscience models, e.g., ring attractor models which has rotational invariance, in general such symmetry does not exist, meaning that the optimal solution is likely unique (up to a reflection). In our experiments, we also show that ground-truth orthogonal transformation can always be recovered, even between nonlinear systems. Nevertheless, while orthogonal transformations possess so many desirable properties, there {obviously exist cases where non-orthogonal transformations are useful}, and we believe the extension of the diffeomorphic alignment framework to non-orthogonal cases is clearly valuable.

\subsection{Nonlinear alignment}

Now we consider the most general case with $f$, $g$ and $\varphi$ all being nonlinear. We have
\begin{flalign}
& (\varphi_{2})_{*}f = (\varphi_{1})_{*}f \\
\iff & (\varphi_{1}^{-1})_{*}(\varphi_{2})_{*}f = f \\
\iff & (\varphi_{1}^{-1}\circ \varphi_{2})_{*}f = f \,,
\end{flalign}
which provides the following theorem:

\begin{Theorem} \label{th:nonlinear}
Given a nonlinear system $\dot{x} = f(x)$ with $f\in C^{\infty}(\mathbb{R}^{n})$ and a diffeomorphism $\varphi_{1} \in \text{Diff}(\mathbb{R}^{n})$, the transformation $\varphi_{2} = \varphi_{1}\circ\psi$ is a diffeomorphism and $(\varphi_{1})_{*}f = (\varphi_{2})_{*}f$ if and only if $\psi = \varphi_{1}^{-1}\circ\varphi_{2}$ is a diffeomorphism and $\psi_{*}f = f$.
\end{Theorem}

This theorem reveals that solution is unique up to the symmetry group of the vector field $f$. There are two cases to consider: First, $f$ may obey some discrete symmetries characterized by discrete groups. For example, if $-f(x) = f(-x)$, then $f$ is invariant under the action of the sign symmetry group $\mathbb{Z}_{2}$ , and we can let $\psi(x) = -x$ or namely $\varphi_{2}(x) = \varphi_{1}(-x)$.
Second, and more interestingly, $f$ might also obey some continuous symmetries characterized by Lie groups. In this case, there will be a continuum of solutions that produce the same pushforward vector field, making it impossible to recover the `ground truth'. One particular example is the one-parameter group $\Phi_{0t}^{f}$ which deforms the vector field $f$ along its own trajectories. Similar to the linear case with $J = e^{TA}$, the transformation $\Phi_{0t}^{f}$ is again a temporal translation, thus not altering the time-invariant vector field $f$.

Due to the existence of this continuous symmetry, the recovery of ground-truth will be impossible unless we apply more constraints. To see this, consider a transformation $\varphi_{1} = \Phi_{01}^{v}$ for some time-varying vector field $v$. {It} is always possible to construct another transformation $\varphi_{2} = \Phi_{01}^{u}$ such that the pushforward vector field $(\varphi_{1})_{*}f = (\varphi_{2})_{*}f$, with
\begin{equation}
    u(t, x) = 
    \begin{cases}
        2f(x)  & t \le 0.5 \\
        2v(2t - 1, x) & t > 0.5
    \end{cases}
\end{equation}
which produces the transformation $\varphi_{2} = \varphi_{1}\circ\Phi_{01}^{f}$.
Therefore, in the DFORM method, we require the diffeomorphisms to be generated by flows of time-invariant vector fields, which significantly mitigates this problem.

\section{Symmetry of the orbital similarity loss} \label{app:symmetry}

In Section~\ref{sec:DFORM}, we mentioned that the orbital similarity loss $J_{f, g, \varphi}(y)$ between $\varphi_{*}f$ and $g$ (Eq.(\ref{eq:topoloss})) is usually numerically different from the loss $J_{g, f, \varphi^{-1}}(x)$ between $(\varphi^{-1})_{*}g$ and $f$ Eq.(\ref{eq:topoloss_inv}). However, there are indeed two special cases where the two terms are related:
First, when $f$ and $g$ are equivalent and their coordinates are related by $\varphi$, obviously both terms vanishes everywhere.
Second, when $\varphi$ is an orthogonal transformation, we have $J_{f, g, \varphi}(\varphi(x)) = J_{g, f, \varphi^{-1}}(x)$. Assuming that $\varphi(x) = Ox$ with $O$ being an orthogonal matrix, we have
\begin{align}
    J_{f, g, \varphi}(\varphi(x)) & = \frac{1}{n}\left\| \frac{Of(x)}{\left\|Of(x)\right\|_2} - \frac{g(Ox)}{\|g(Ox)\|_{2}} \right\|_2^2 \\
    & = \frac{1}{n}\left\| \frac{Of(x)}{\left\|f(x)\right\|_2} - \frac{g(Ox)}{\|O^{-1}g(Ox)\|_{2}} \right\|_2^2 \\
    & = \frac{1}{n}\left\| O\left(\frac{f(x)}{\left\|f(x)\right\|_2} - \frac{O^{-1}g(Ox)}{\|O^{-1}g(Ox)\|_{2}}\right) \right\|_2^2 \\
    & = \frac{1}{n}\left\| \frac{f(x)}{\left\|f(x)\right\|_2} - \frac{O^{-1}g(Ox)}{\|O^{-1}g(Ox)\|_{2}} \right\|_2^2 \\
    & = J_{g, f, \varphi^{-1}}(x)\,.
\end{align}

A similar conclusion holds for the orbital similarity index. If $f$ and $g$ are equivalent and their coordinates are related by $\varphi$, then the orbital similarity is one for all $x$ or $y$; if $\varphi$ is an orthogonal transformation, then $\cos\angle\left( \varphi_{*}f(\varphi(x)), g(\varphi(x)) \right)$ equals $\cos\angle\left(f(x), (\varphi^{-1})_{*}g(x)\right)$.

\clearpage

\section{Hyperparameters} \label{app:hyperparameter}

\subsection{DFORM default hyperparameters}

\begin{table}[h!]
\renewcommand{\arraystretch}{1.3}
\caption{Hyperparameters for loss calculation.}
\begin{center}
\begin{tabular}{|c|c|c|}
\hline
    Notation & Default value & Explanation \\
\hline
    $\lambda_{1}$ & 1 & Weights for loss term $l_{1}$ \\
    $\lambda_{2}$ & 1 & Weights for loss term $l_{2}$ \\
    $\lambda_{3}$ & 0 & Weights for loss term $l_{3}$ \\
    $\lambda_{4}$ & 0 & Weights for loss term $l_{4}$ \\
    $\lambda_{5}$ & 0 & Weights for regularization term $\frac{1}{n}\|v\|_{H}^{2}$ \\
    $\lambda_{6}$ & 0 & Weights for regularization term $\frac{1}{n}\|HH^{T} - I_{n}\|_{F}^{2}$ \\
    \verb|f_samp| & $\mathcal{N}(\mathbf{0}_{n}, I_{n})$ & {Sample} distribution for system $f$ \\
    \verb|g_samp| & $\mathcal{N}(\mathbf{0}_{n}, I_{n})$ & {Sample} distribution for system $g$ \\
    \verb|warp_time| & True & Whether to normalize the norm of vector fields in loss \\
\hline
\end{tabular}
\end{center}
\end{table}

\begin{table}[hp]
\renewcommand{\arraystretch}{1.3}
\caption{Hyperparameters for model architecture.}
\begin{center}
\begin{tabular}{ | m{10em} | m{6em} | m{20em} | }
\hline
    Notation & Default value & Explanation \\
\hline
    \verb|add_linear| & True & Whether to include linear layer \\
    \verb|time_varying| & False & Whether to make the Neural ODE time-dependent \\
    \verb|n_hid| & [$\max\{2n, 20\}$, $\max\{2n, 20\}$] & Width of each hidden layer in Neural ODE \\
    \verb|act_fn| & ELU & Activation function for Neural ODE \\
    \verb|phi_samp| & $\mathcal{N}(\mathbf{0}_{n}, I_{n})$ & Sampling distribution to calculate $\|v\|_{H}$ \\
    \verb|phi_samp_size| & 128 & Number of samples used to calculate $\|v\|_{H}$ \\
    \verb|id_init_nonlinear| & True & Whether to initialize the Neural ODE to be the identity function \\
\hline
\end{tabular}
\end{center}
\end{table}

\begin{table}[hp]
\renewcommand{\arraystretch}{1.3}
\caption{Hyperparameters for optimization.}
\begin{center}
\begin{tabular}{|c|c|c|}
\hline
    Notation & Default value & Explanation \\
\hline
    \verb|batch_size| & 128 & Number of random $x$ and $y$ to drawn in each batch \\
    \verb|nBatch_linear| & NA& Number of batches for training the linear layer \\
    \verb|nBatch| & NA& Number of batches for full-scale training \\
    \verb|lr_linear| & 0.002& Learning rate for training the linear layer \\
    \verb|lr| & 0.002& Learning rate for full-scale training \\
    \verb|n_rep| & 1 & Number of repetitions \\
\hline
\end{tabular}
\end{center}
\end{table}

\clearpage

\subsection{Specific hyperparameters for each experiment}

\begin{table}[hp]
\renewcommand{\arraystretch}{1.3}
\caption{Hyperparameters for loss calculation and architecture for each experiment.}
\begin{center}
\begin{tabular}{|c|c|c|c|c|c|c|c|c|c|}
\hline
    Section(s) & $\lambda_{1}$ & $\lambda_{2}$ & $\lambda_{3}$ & $\lambda_{4}$ & $\lambda_{5}$ & $\lambda_{6}$ & \verb|f_samp| & \verb|g_samp| & \verb|add_linear|\\
\hline
    \ref{sec:linear_equiv}, \ref{sec:linear_sgndiff} & 1 & 1 & 0 & 0 & NA & 0.001 & Gaussian & Gaussian & True \\
    \ref{sec:equiv_RNN} & 1 & 1 & 0 & 0 & NA & 0.001 & Gaussian & Pushforward & True \\
    \ref{sec:equiv_VDP} & 1 & 1 & 0 & 0 & 0 & 0 & VDP & VDP & True \\
    \ref{sec:paramrec} {A} & {1} & {1} & {0} & {0} & {0.001} & {0.1} & {Gaussian} & {Pushforward} & {True/False}\\
    \ref{sec:paramrec} {B-C} & 1 & 1 & 0 & 0 & $10^{-5}$ & NA & Gaussian & Pushforward & {True/}False \\
    \ref{sec:BLA} & {1} & {1} & {0} & {0} & {0.1} & {0.1} & {Uniform} & {Uniform} & {True}\\
    \ref{sec:SNIC} & 1 & 1 & 10 & 10 & NA & 0.001 & Pushforward & SNIC & True \\
    \ref{sec:saddle-cycle} & 1 & 1 & 10 & 10 & NA & 0.1 & Pushforward & Hopf & True \\
    \ref{sec:MINDy_template} & 1 & 10 & 10 & 100 & NA & 0.001 & Asymptotic & Asymptotic & True \\
\hline
\end{tabular}
\end{center}
\end{table}

\begin{table}[hp]
\renewcommand{\arraystretch}{1.3}
\caption{Hyperparameters for architecture and training for each experiment.}
\begin{center}
\begin{tabular}{|c|c|c|c|c|c|c|}
\hline
    Section(s) & \verb|batch_size| & \verb|nBatch_linear| & \verb|nBatch| & \verb|lr_linear| & \verb|lr| & \verb|n_rep| \\
    \ref{sec:linear_equiv}, \ref{sec:linear_sgndiff} & 128& 2500& NA& 0.002& NA & {3}\\
    \ref{sec:equiv_RNN} & 32& 20000& NA& 0.002& NA & 5\\
    \ref{sec:equiv_VDP} & 32& 2000/5000& 3000/0& 0.002& 0.0002& 1\\
    \ref{sec:paramrec} {A} & {32} & {0/2000} & {2000/0} & {0.001} & {0.001} & {1}\\
    \ref{sec:paramrec} {B-C} & 32& {0/10000} & {10000/0} & {0.001}& 0.001& 5\\
    \ref{sec:BLA} & {32} & {2000} & {3000} & {0.002} & {0.0002} & {5}\\
    \ref{sec:SNIC} & 32& 20000& NA& 0.002& NA & 1\\
    \ref{sec:saddle-cycle} & 32& 3000& NA& 0.002& NA & 1\\
    \ref{sec:MINDy_template} & 128& 3000& NA& 0.002& NA & 2\\    
\hline
\end{tabular}
\end{center}
\end{table}

\subsection{{Robustness against hyperparameter variations}}

\paragraph{{Loss weights $\lambda_{1}$ to $\lambda_{4}$.}}
{In Sections~\ref{app:sampling} and \ref{app:crossdim}, we have discussed how to choose loss term weights $\lambda_{1}$ to $\lambda_{4}$ according to questions of interest. Here we show that when the two systems are indeed topologically equivalent, the results were quite robust against any combination of weights. We used the same Van der Pol oscillator systems from Section~\ref{sec:equiv_VDP} and swept through the hyperparameter space while requiring that $\lambda_{2} : \lambda_{1} = \lambda_{4} : \lambda_{3}$ and $\lambda_{3} : \lambda_{1} = \lambda_{4} : \lambda_{2}$. The first ratio $r_{1}$ quantifies the relative importance of the quality of the inverse transformation $\varphi^{-1}$ compared to that of $\varphi$, and it was set to one by default. The second ratio $r_{2}$ quantifies the relative importance of the loss calculated by averaging across the pushforward measure compared to across the target measure, and was zero by default. We performed a grid search over the combinations of these two ratios at seven levels each: $[0, 0.01, 0.1, 1, 10, 100, \infty]$. For example, $(r_{1}, r_{2}) = (0.1, 100)$ corresponds to $(\lambda_{1}, \lambda_{2}, \lambda_{3}, \lambda_{4}) = (1, 0.1, 100, 10)$, while $(r_{1}, r_{2}) = (\infty, \infty)$ corresponds to $(\lambda_{1}, \lambda_{2}, \lambda_{3}, \lambda_{4}) = (0, 0, 0, 1)$. Each experiment was repeated with three different random initializations and the one with the best orbital similarity was retained. Other hyperparameters were the same as in Section~\ref{sec:equiv_VDP}. In particular, $\lambda_{5} = \lambda_{6} = 0$. We visualized the learned pushforward vector fields and distributions in the same coordinate system in Figure~\ref{fig:g2fW-domainW}, using the same convention as in the main Figure~\ref{fig:VDP}. One can see that while the hyperparameters varied dramatically, the learned pushforwards were in general quite similar.}

\paragraph{{Regularization weights $\lambda_{5}$ and $\lambda_{6}$.}}
{In most experiments in the main text, we applied none or very weak regularization. However, in many cases, a strong regularization might be necessary to ensure that the learned mapping is reasonable. To illustrate the effect of different levels of regularization, we again adopted the Van der Pol oscillator example in Section~\ref{sec:equiv_VDP} and swept through combinations of hyperparameters $\lambda_{5}$ and $\lambda_{6}$ each at six different levels: $[0, 0.01, 0.1, 1, 10, 100]$. Loss weights were still set to $(\lambda_{1}, \lambda_{2}, \lambda_{3}, \lambda_{4}) = (1, 1, 0, 0)$ and other hyperparameters also remained unchanged. The model with the best orbital similarity was selected out of three random initializations. Learned pushforward vector fields and distributions were visualized in Figure~\ref{fig:regularization}. The results were fairly similar when $\lambda_{5}, \lambda_{6}\le 1$. As $\lambda_{5}$ further increased, the solution became more and more linear. The effect of increasing $\lambda_{6}$ was small when $\lambda_{5}$ was small, as the linear transformation can be compensated by the nonlinear flow. When both terms were big, the solution became close to orthogonal.}

\paragraph{{Batch size.}}
{To investigate the influence of different training batch sizes on the learned solution, we repeated the linear alignment experiment on non-orthogonally transformed 32-dimensional RNNs in Section~\ref{sec:equiv_RNN}, with batch size 16 and 64 instead of the original 32. Results across all repetitions were visualized in Figure~\ref{fig:batchsize}. The distribution of learned fixed point and Jacobian alignment did not show a sizable difference between the three choices of batch size.}

\paragraph{{Sample distribution.}} \label{app:distri}
{The sample distributions $p_{x}$ and $p_{y}$ play an important role in DFORM training. In the main text, we mainly adopted two strategies: 1) an isotropic normal distribution or an uniform distribution over some regions-of-interest; or 2) the asymptotic distribution of states when simulating the system under realistic noise. Here, we explore how the solution changes as we interpolate between these two types of distributions. We repeated the MINDy template matching experiments in Section~\ref{sec:MINDy_template} with three different mixtures of the asymptotic distribution (as in the main text) and the isotropic standard normal distribution. The probability of sampling from the latter increased from 5\% to 15\%, then 50\%, essentially weighing the regions near the origin more and more compared to the stable limit sets. We visualized the vector field similarity and learned pushforward for each case in Figure~\ref{fig:distri} using the same convention as in Figure~\ref{fig:MINDy}. As the weight for isotropic normal distribution increased, the quality of matching for the limit cycle decreased, so was the difference between similarity scores in the limit cycle models and other models. Therefore, when the question of interest is to compare two models around the asymptotically stable regions, it would be important to adopt a measure that mostly covers these regions.}

\clearpage

\section{Supplementary Figures}

\renewcommand{\thefigure}{S\arabic{figure}}
\setcounter{figure}{0}

\begin{figure}[htbp]

\begin{center}
\includegraphics[width=0.7\textwidth]{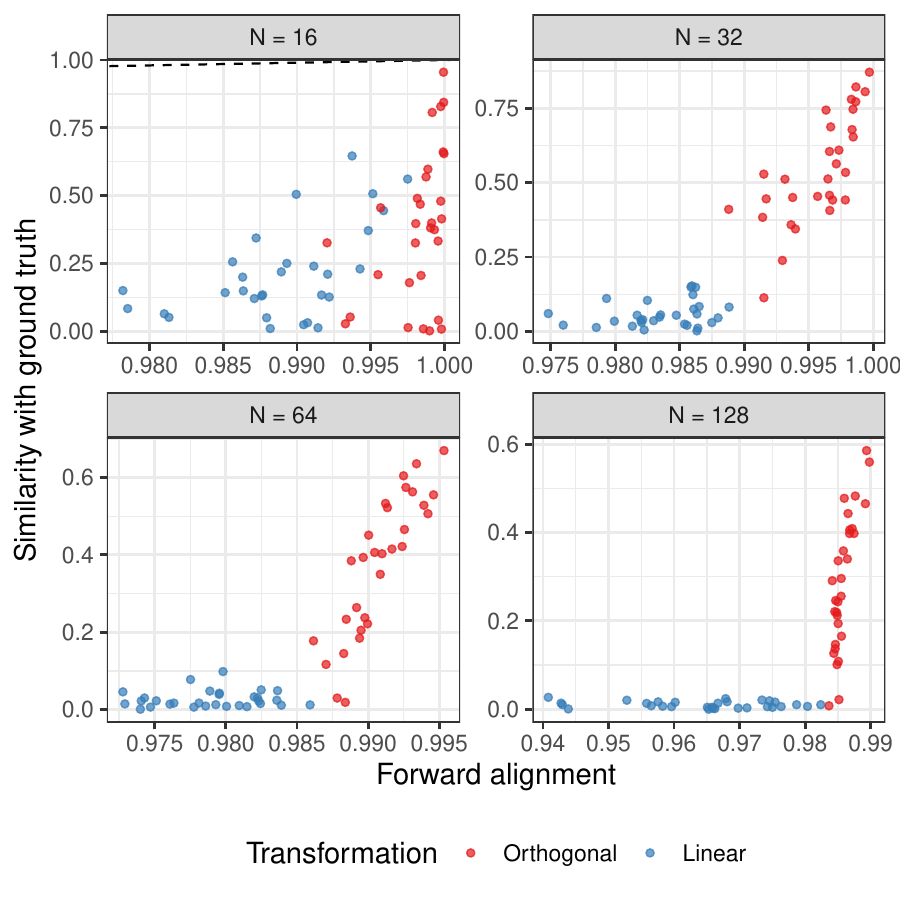}
\end{center}
\caption{{\bf Alignment between diffeomorphic linear systems.} X axis represents the vector field alignment between transformed and target systems after DFORM training. Y axis represents the cosine similarity between the linear transformation matrix learned by DFORM and the ground truth matrix. Each dot indicates one experiment, with color representing the type of ground truth transformation. Different panels showed results for systems of different sizes.}
\label{fig:linear-scatter}
\end{figure}

\begin{figure}[htbp]

\begin{center}
\includegraphics[width=\textwidth]{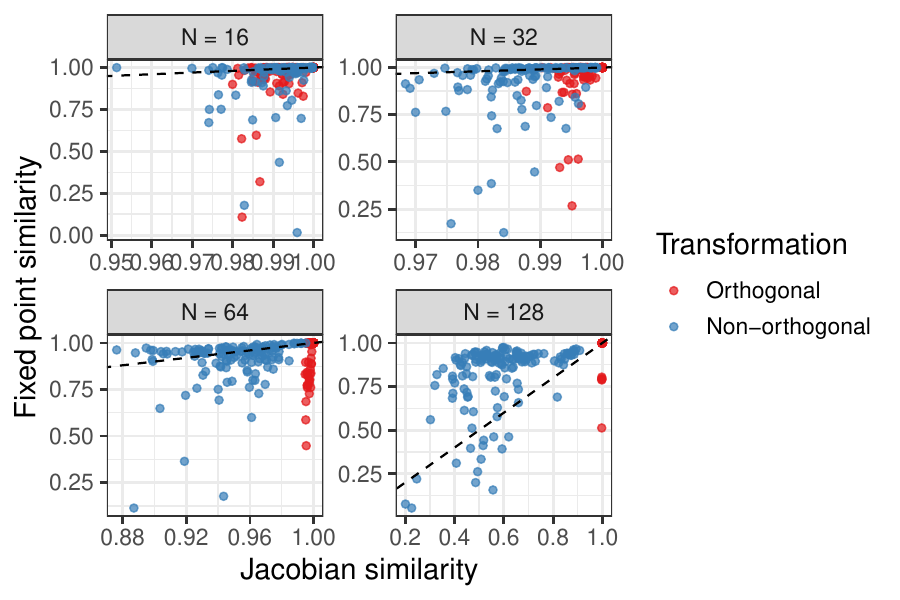}
\end{center}
\caption{{{\bf Alignment between linearly-transformed RNNs.} Similar convention with main Figure~\ref{fig:RNN}, but showing the results from all repetitions instead of only the ones with the highest fixed point alignment.}}
\label{fig:RNN-scatter-all}
\end{figure}

\begin{figure}[htbp]

\begin{center}
\includegraphics[width=\textwidth]{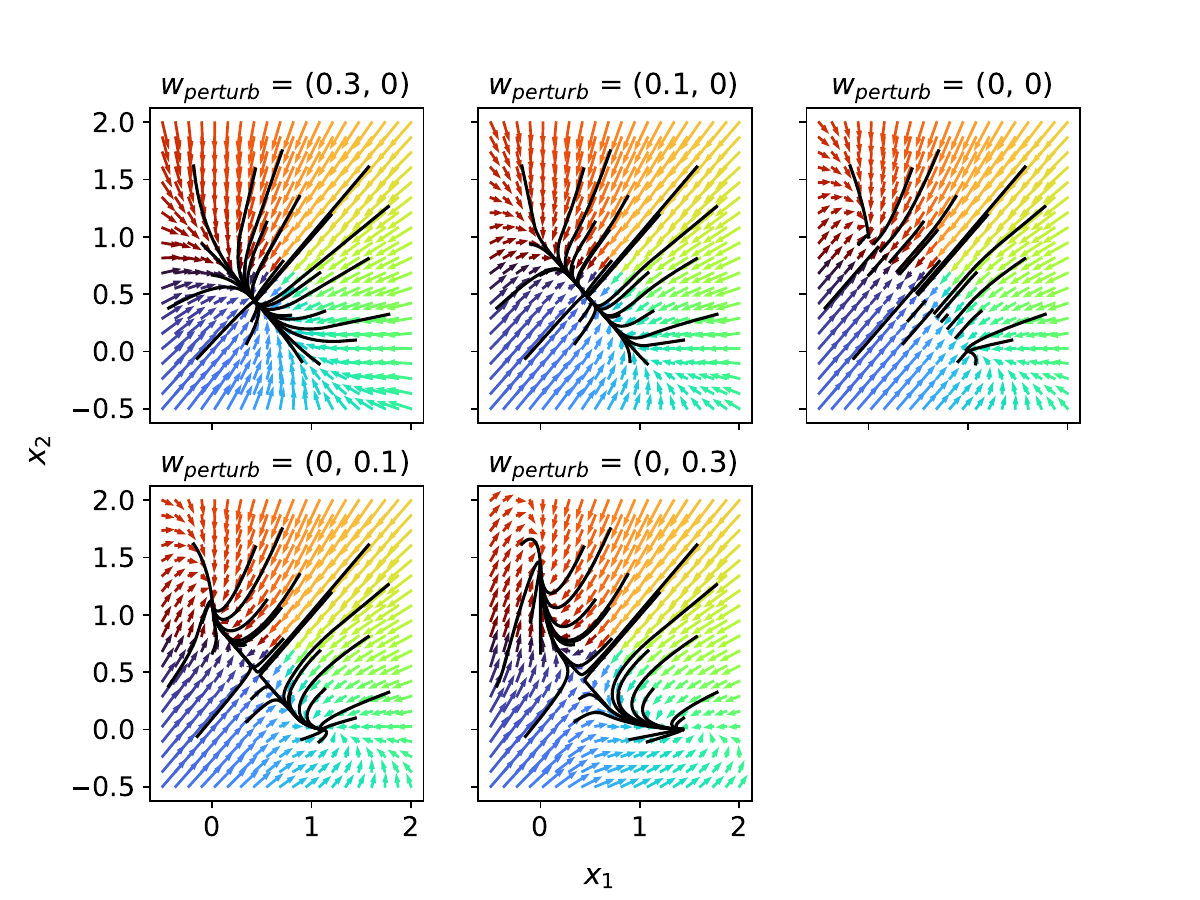}
\end{center}
\caption{{{\bf Bounded line attractor system and its perturbations.} Black traces showed simulated trajectories. Quiver plots showed vector fields.}}
\label{fig:BLA-sys}
\end{figure}

\begin{figure}[htbp]

\begin{center}
\includegraphics[width=\textwidth]{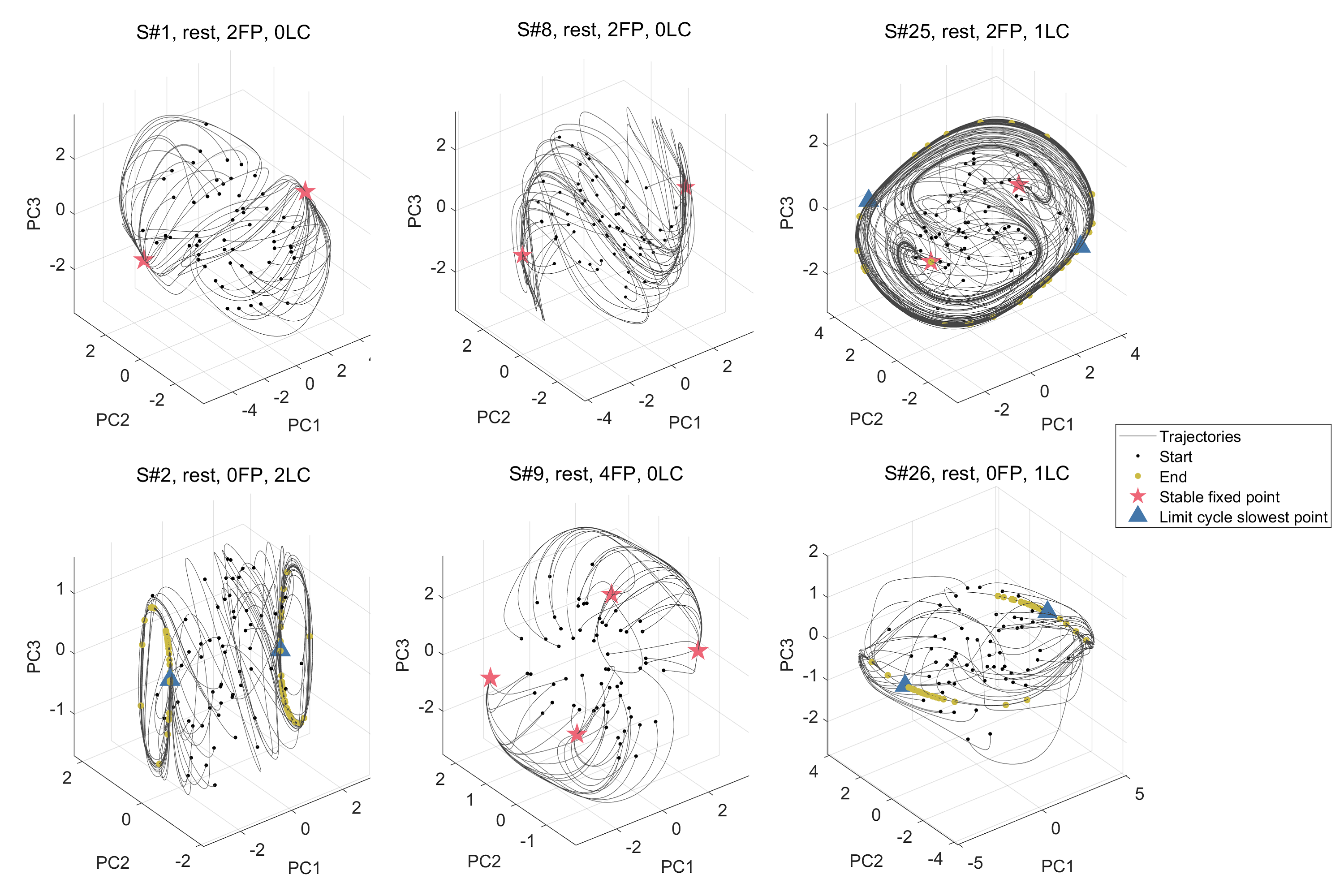}
\end{center}
\caption{{{\bf Trajectories and attractors of example MINDy models.} 60 trajectories with random initial conditions were simulated and projected to the first three principal components (PCs). Black and yellow dots represent starting and end point of the trajectories. Red star indicates numerically identified stable fixed point. Blue triangle indicates numerically identified slowest points on limit cycles. Title indicates the model index, data source (here `rest', resting state fMRI), number of stable fixed points (FP) and limit cycles (LC). Out of all 30 models, 17 have 2FP, 0LC; 9 have 0FP, 1LC; 2 have 4FP, 0LC; 1 has 0FP, 2LC; 1 has 2FP, 1LC. For more information, see \parencite{chenDynamicalModelsReveal2025}.}}
\label{fig:MINDy-traj}
\end{figure}

\begin{figure}[htbp]

\begin{center}
\includegraphics[width=\textwidth]{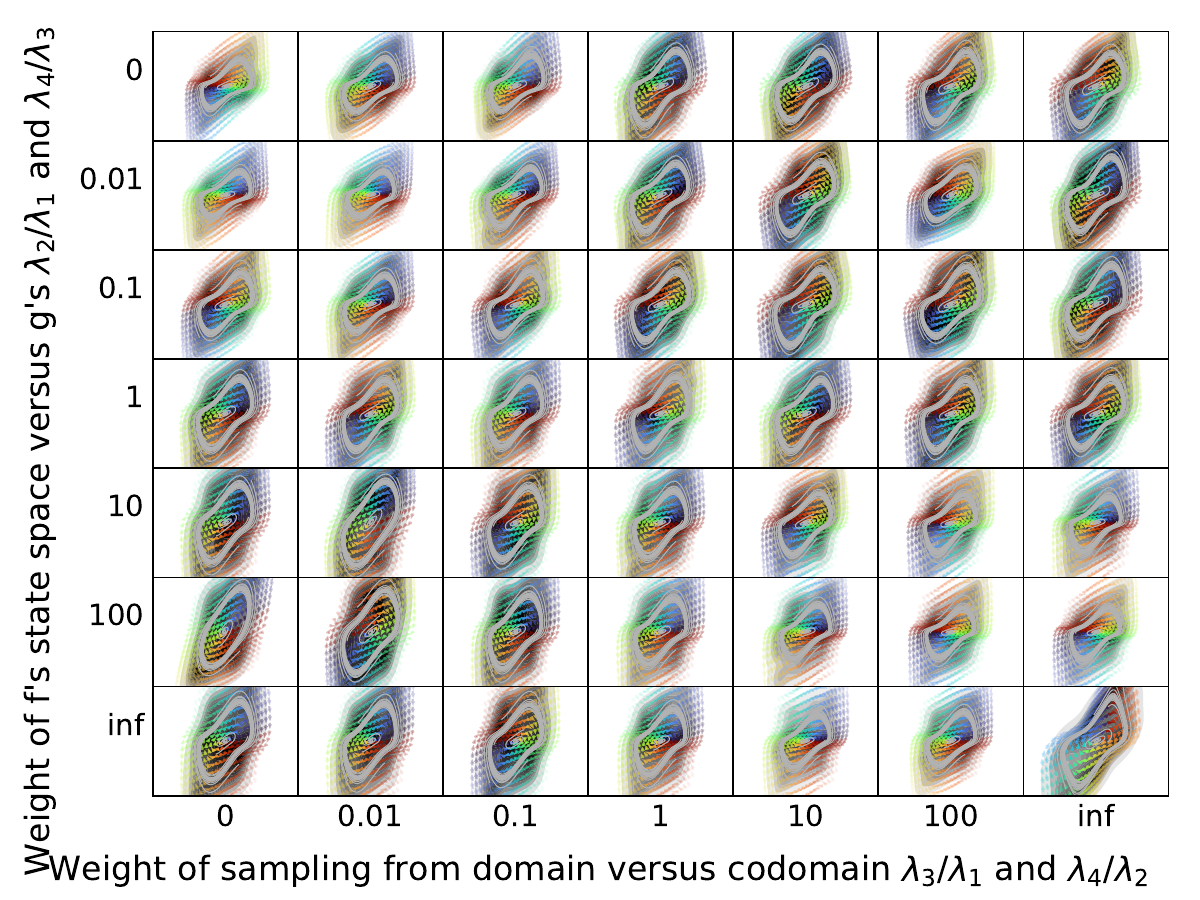}
\end{center}
\caption{{{\bf Robustness against variations of loss term weights.} Learned pushforward vector fields and distributions were visualized in the same convention as the middle panels in Figure~\ref{fig:VDP}. Panels were arranged according to the ratio $\lambda_{2} : \lambda_{1}$ and $\lambda_{4} : \lambda_{3}$ (rows) and the ratio $\lambda_{3} : \lambda_{1}$ and $\lambda_{4} : \lambda_{2}$ (columns). Results in Figure~\ref{fig:VDP} was obtained using the same configuration as the first panel on the fourth row.}}
\label{fig:g2fW-domainW}
\end{figure}

\begin{figure}[htbp]

\begin{center}
\includegraphics[width=\textwidth]{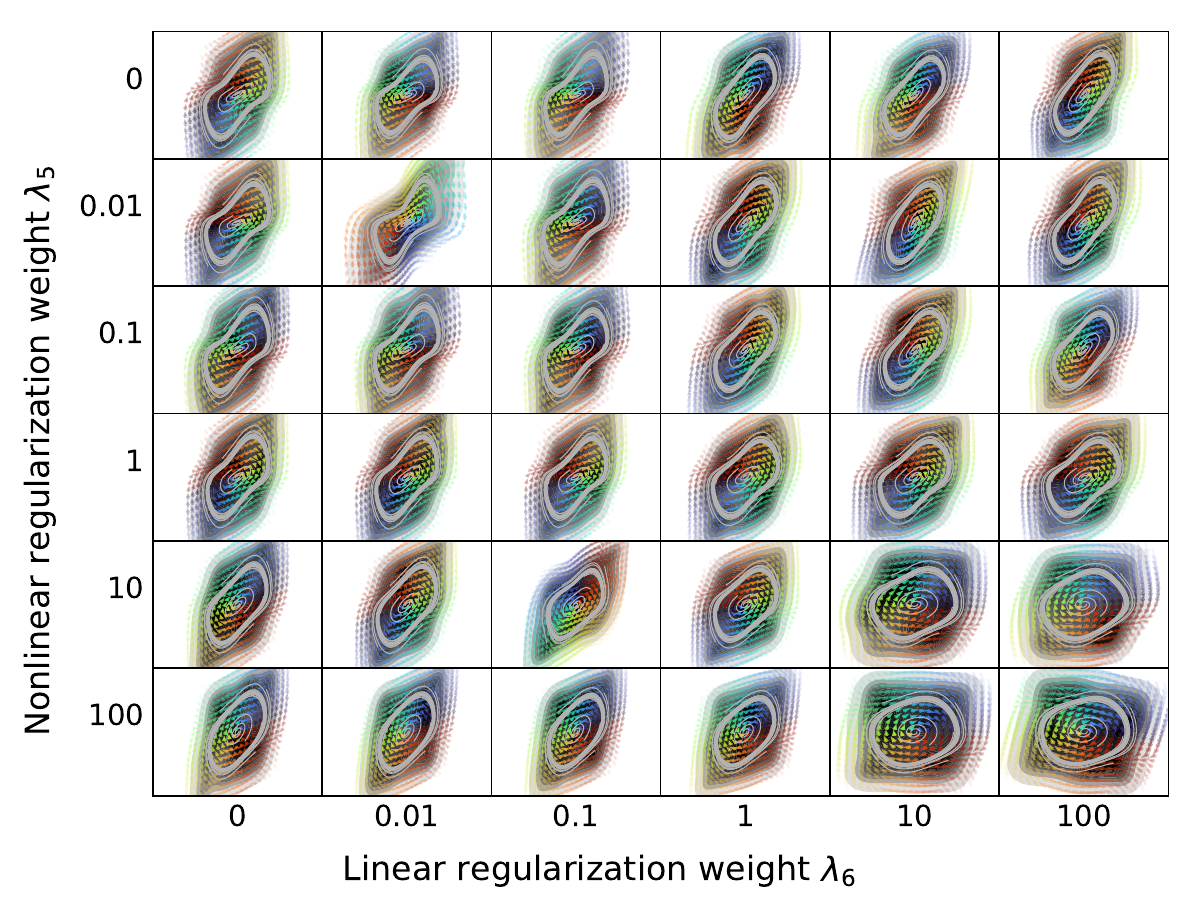}
\end{center}
\caption{{{\bf Robustness against variations of regularization weights.} Learned pushforward vector fields and distributions were visualized in the same convention as the middle panels in Figure~\ref{fig:VDP}. Panels were arranged according to the nonlinear regularization weight $\lambda_{5}$ (rows) and linear orthonormality regularization weights $\lambda_{6}$ (columns). Results in Figure~\ref{fig:VDP} was obtained using the same configuration as the top left panel.}}
\label{fig:regularization}
\end{figure}

\begin{figure}[htbp]

\begin{center}
\includegraphics[width=\textwidth]{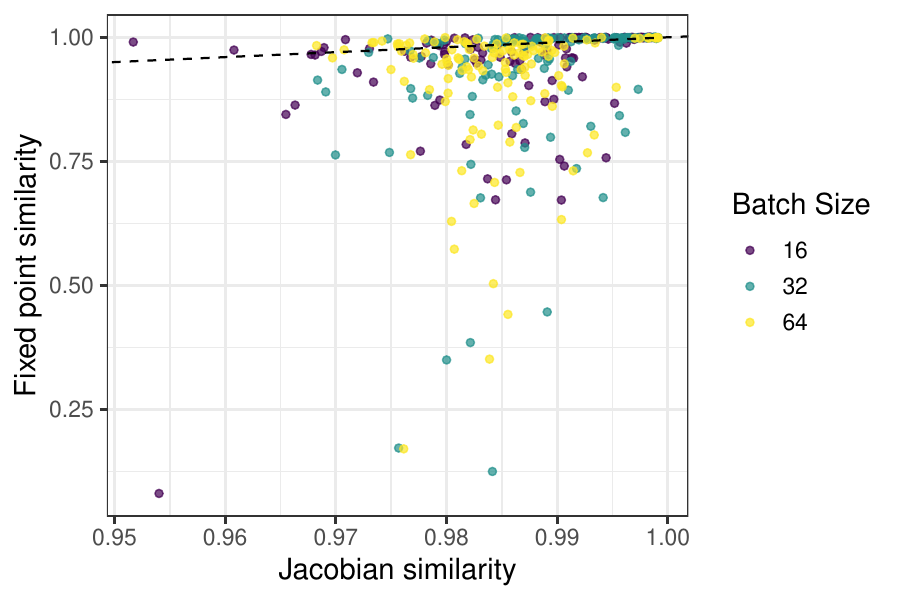}
\end{center}
\caption{{{\bf Robustness against variations of batch size.} Experiments on 32-dimensional non-orthogonally transformed RNNs were repeated as in Section~\ref{sec:equiv_RNN} but with different batch sizes. X and Y coordinates denote the Jacobian and fixed point alignment after training in each experiment. All five repetitions were included.}}
\label{fig:batchsize}
\end{figure}

\begin{figure}[htbp]

\begin{center}
\includegraphics[width=0.95\textwidth]{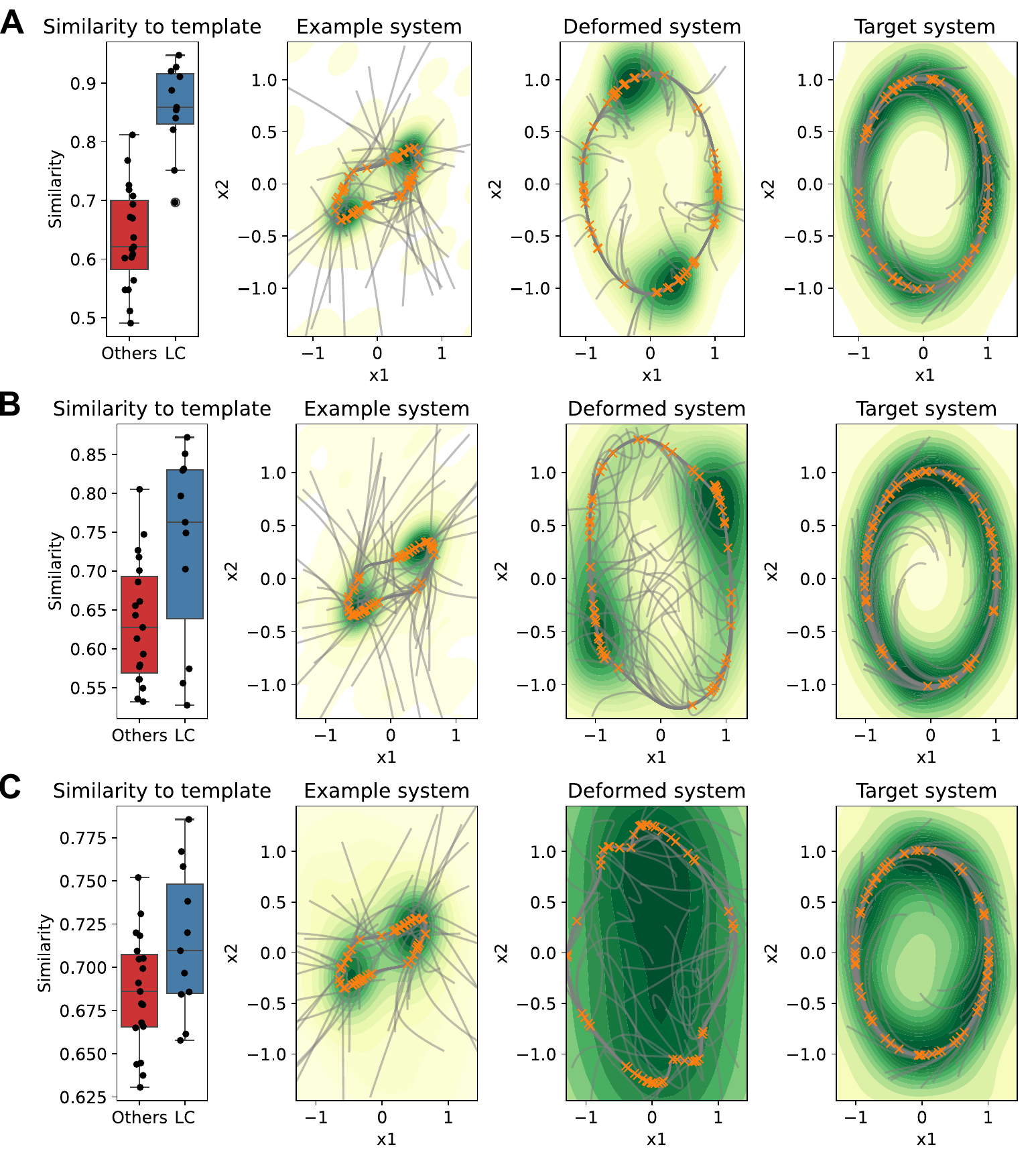}
\end{center}
\caption{{{\bf Effect of different sample distributions.} Experiments on MINDy models in Section~\ref{sec:MINDy_template} were repeated with three different choices of sample distributions. A. 95\% chance from the asymptotic distribution (with noise standard deviation 0.05, as in the main text) and 5\% chance from the isotropic standard normal distribution. B. 85\% chance from asymptotic distribution and 15\% chance from isotropic normal distribution. C. 50\% chance from each. Visualization convention is the same as in Figure~\ref{fig:MINDy}.}}
\label{fig:distri}
\end{figure}

\end{appendices}

\clearpage

\printbibliography

\end{document}